\documentclass[11pt]{article}

\usepackage[final]{acl}

\usepackage{latexsym}

\usepackage{microtype}

\usepackage{inconsolata}

\usepackage{graphicx}

\usepackage{tabularx}
\hypersetup{colorlinks=true, linkcolor=blue, citecolor=blue, urlcolor=blue}
\title{Instructions for *ACL Proceedings}

\usepackage{subcaption}

\author{First Author \\
  Affiliation / Address line 1 \\
  Affiliation / Address line 2 \\
  Affiliation / Address line 3 \\
  \texttt{email@domain} \\\And
  Second Author \\
  Affiliation / Address line 1 \\
  Affiliation / Address line 2 \\
  Affiliation / Address line 3 \\
  \texttt{email@domain} \\}

\title{How Far Do Auto-Interpretation Labels Generalize: A Controlled Study Across Languages, Scripts, and Rewordings}

\author{Sripad Karne \\
  Columbia University \\
  \texttt{sk5695@columbia.edu}}

\usepackage{fontspec}
\newfontfamily\cyrillicfont{FreeSerif}[
  Extension      = .otf,
  UprightFont    = *,
  BoldFont       = *Bold,
  ItalicFont     = *Italic,
  BoldItalicFont = *BoldItalic,
]
\usepackage{ucharclasses}
\setTransitionsForCyrillics{\cyrillicfont}{\normalfont}
\usepackage{booktabs}
\usepackage{float}

\usepackage{multirow}
\usepackage{pifont}   
\usepackage{xcolor}
\usepackage{dblfloatfix}

\usepackage{float}

\begin{document}
\maketitle
\begin{abstract}
Sparse autoencoder (SAE) features are increasingly used to interpret language
models, with auto-generated natural-language labels serving as the primary
interface for understanding what each feature represents. We ask whether these
labels generalize: does a feature labeled for a concept actually track that
concept across languages and scripts? Using Serbian digraphia as a controlled
testbed---the same language written in both Latin and Cyrillic via
deterministic transliteration---we first find that SAE feature sets activated by the same content in
different languages, scripts, and wordings share substantial overlap (mean Jaccard 0.39 vs.\ 0.13 random baseline, peaking at 0.57), suggesting genuine
cross-lingual semantic features. We then test whether
auto-interpretation labels keep pace. They often do not: features whose labels describe
semantic content miss the same meaning in Serbian up to $4\times$ more often
than within English, and miss Serbian Cyrillic more than
Serbian Latin---two scripts that are deterministic transliterations of each
other. The gap grows with network depth, yet the labels give no indication
that they fail.
These results suggest that auto-interpretation labels reflect a feature's behavior on the languages and scripts a model has seen most in training, rather than the concept itself.
\end{abstract}

\section{Introduction}
\label{sec:intro}

Sparse autoencoders (SAEs) have become a standard tool for inspecting the
internals of language models, decomposing dense activations into sparse
features that are easier to interpret
\citep{bricken2023, cunningham2024, templeton2024}. Because models contain far
more features than anyone can inspect by hand, each feature is typically
labeled automatically: a language model reads the feature's top-activating
examples and writes a short natural-language description
\citep{bills2023language, neuronpedia}. These labels are then what practitioners, such as interpretability researchers
and auditors using feature directories like Neuronpedia, rely on---to understand a model, to audit its behavior, or to
steer it.

But how far do these labels generalize? Consider a feature labeled
``deception'' or ``violent content.'' A researcher reads the label, takes the
feature to track that concept, and relies on it---perhaps to monitor safety-relevant
behavior. What the label does not say is whether the feature tracks that concept
equally well across languages and scripts.
If a feature detects deception mainly in English, weakening in Russian and
weakening further in Serbian Cyrillic, then the same content that triggers it
in one form could pass unnoticed in another. The label may name the concept
correctly; what it omits is the \emph{range} over which the feature actually
tracks it.

Testing this requires a setting where surface form can be varied while
meaning is held exactly fixed. Serbian digraphia offers precisely this
control: Serbian is
natively written in both Latin and Cyrillic scripts, related by deterministic,
lossless transliteration, so we can change script while holding language,
wording, and meaning constant. Combined with cross-language controls
(English, Russian), we construct a factorial paradigm that isolates script,
language, wording, and meaning independently.

First, in what we call Leg~1 of our analysis, we find evidence that SAE features encode abstract meaning: feature
sets activated by the same content across different languages, scripts, and
wordings overlap substantially (mean Jaccard 0.39 vs.\ 0.13 baseline, peaking at 0.57), a
pattern robust across model scales, architectures, and SAE hyperparameters, providing evidence that cross-lingual semantic features exist and are worth
labeling. Second, in Leg~2, we test whether the auto-interpretation labels
assigned to these features hold up when the same content is rendered in
Serbian. They do not. Content-labeled features miss the same meaning in Serbian up to $4\times$
more often than within English---a gap that grows with network depth and that
the labels themselves give no indication of---and miss Serbian Cyrillic up to $1.35\times$ more than Serbian Latin despite the two being identical up to transliteration.
Our contributions are:
\begin{enumerate}
\item A controlled cross-lingual evaluation showing that auto-interpretation
labels can fail systematically on less-represented languages and scripts,
with failures aligning with estimated training coverage, and that this
failure is invisible from the label itself.
\item A factorial paradigm built on Serbian digraphia, two native scripts of one language related by lossless transliteration, that isolates script, language, wording, and meaning in SAE feature sets, confirming cross-lingual semantic features at the feature-set level.
\item A controlled multilingual evaluation suite built on FLORES+: 300
sentences in four language-script variants with validated paraphrases and
matched random partners, designed for factorial analysis of feature-level interpretability methods and
released to support future work.\footnote{Dataset and code available at
\url{https://github.com/Sripadkarne/auto-interp-cross-lingual-eval}.}
\end{enumerate}

\section{Related Work}
\label{sec:related}

\paragraph{Sparse autoencoders and feature interpretation.}
Sparse autoencoders (SAEs) decompose model activations into sparse, more
interpretable features, offering a route to inspecting representations beyond
individual neurons \citep{bricken2023, cunningham2024, gao2024scaling}. Open SAE suites such as Gemma Scope 2 have made this
practical at scale \citep{gemmascope2_2025}, and feature directories built on them are now a
common entry point for interpretability work. We use Gemma Scope 2 SAEs and the labels
served through one such directory, treating them as the deployed tools a practitioner
would actually reach for.

\paragraph{Automated interpretation and its reliability.}
Features are typically labeled automatically through auto-interpretation
\citep{bills2023language, neuronpedia}. Later pipelines have built on this approach: Delphi
\citep{paulo2025automatically} scales labeling to millions of features, and PRISM \citep{kopf2025prism} extends it to multiple descriptions
per feature to capture polysemanticity. Across these pipelines, however, the reliability of the resulting labels remains
an open concern. Prior work shows that
automated explanations can be vague or inaccurate, and has proposed more rigorous ways to
assess them \citep{huang2023rigorously, liu2026neuronscope}. These evaluations,
however, are monolingual: they ask whether a label fits a feature's behavior on the inputs
it was derived from, not whether it still holds when the same content appears in another
language or script. To our knowledge, that gap has not been tested, and it is what we test here.

\paragraph{Multilingual representations and script.}
Multilingual models are known to represent meaning in a partly
language-agnostic way: equivalent sentences in different languages converge
in a shared semantic space in the middle layers, before representations
specialize toward the output language near the end \citep{wendler2024,
wu2025semantichub}. This shared structure has been probed at the SAE feature level, where studies recover language-specific features \citep{deng2025unveiling} and grammatical concepts shared across typologically diverse languages \citep{brinkmann2025large}. \citet{verma2026} found that language-associated features are strongly conditioned on script: romanizing a non-Latin language activates a largely disjoint set of features from its native form. Romanized text, however, is a largely synthetic form a model rarely encounters, so it changes script and naturalness at once. Serbian digraphia instead contrasts two scripts the language is genuinely written in, isolating script within forms the model has actually seen. This control is what lets us test feature labels cleanly: if features track meaning across languages and scripts, the labels assigned to them should hold across them too. We test whether they do.

\section{Methods}
\subsection{Dataset}

\paragraph{Source corpus.} We draw 300 sentences from the FLORES+ devtest split \citep{goyal-etal-2022-flores, nllb-2024, maillard-etal-2024-findings}, a multilingual benchmark of professionally translated sentences, stratified across its Wikinews, Wikijunior, and Wikivoyage sources (roughly 100 each) to span news, instructional, and travel registers (Appendix~\ref{app:domain}). FLORES+ supplies aligned professional translations of each sentence into Serbian (Cyrillic) and Russian (Cyrillic).

\paragraph{Language and script variants.} Each sentence appears in four language-script variants: English in Latin, Serbian in Cyrillic, Serbian in Latin, and Russian in Cyrillic. Serbian Latin is produced by deterministic, lossless transliteration of Serbian Cyrillic using the Vuk Karadžić mapping implemented in implemented in \texttt{cyrtranslit}\footnote{\url{https://zenodo.org/records/17663256}}. The two Serbian variants thus differ only in script while holding content fixed, the property our paradigm exploits to isolate script from language and meaning.

\paragraph{Conditions.} Within each variant, every sentence appears in three conditions. The \emph{original} is the FLORES+ professional translation, or its transliteration for Serbian Latin. The \emph{paraphrase} is a meaning-preserving rewrite with varied surface form, generated and validated as described below. The \emph{random partner} is an controlled-unrelated FLORES+ sentence in the same language and script, length-matched to within three words; it shares script and language with the target but no semantic content. Table~\ref{tab:design} summarizes the full set of variants and conditions. Crossing the four language-script variants with the three conditions (original, paraphrase, random partner) across all 300 sentences yields $300 \times 4 \times 3 = 3{,}600$ texts.

\begin{table}[t]
\centering
\small
\begin{tabular}{@{}lll@{}}
\toprule
\textbf{Variant} & \textbf{Language} & \textbf{Script} \\
\midrule
En-Latin    & English & Latin \\
Sr-Latin    & Serbian & Latin \\
Sr-Cyrillic & Serbian & Cyrillic \\
Ru-Cyrillic & Russian & Cyrillic \\
\midrule
\textbf{Condition} & \multicolumn{2}{l}{\textbf{Description}} \\
\midrule
Original   & \multicolumn{2}{l}{FLORES+ reference translation} \\
Paraphrase & \multicolumn{2}{l}{Meaning-preserving rewrite} \\
Random     & \multicolumn{2}{l}{Controlled-unrelated sentence, same variant} \\
\bottomrule
\end{tabular}
\caption{The four language-script variants and three conditions.}
\label{tab:design}
\end{table}

\paragraph{Paraphrase generation.} English paraphrases were generated with
Claude Opus 4.6, prompted to preserve meaning while varying surface form
and matching length within 30\% from the original FLORES+ phrase. Serbian and Russian paraphrases were
obtained by translating the English paraphrase with the same model; Serbian
Latin by transliterating Serbian Cyrillic. All candidates were filtered with
LaBSE \citep{feng-etal-2022-language} (cosine similarity $\geq 0.80$). Full
prompts appear in Appendix~\ref{app:prompts}.

\paragraph{Native-speaker validation.} For each of English, Serbian, and Russian we recruited two validators. Each judged 200 sentences with a 100-sentence overlap between the pair, covering 300 unique sentences per language. Each sentence pair received two binary judgments: whether the paraphrase preserves the original \emph{meaning}, and whether it reads as \emph{natural} text. A pair passing both was accepted as is; otherwise the validator supplied a corrected paraphrase, which was added to the dataset and flagged for reference (full instructions in Appendix~\ref{app:validation}). These corrections do not drive our  results, with no robust difference between
accepted and corrected paraphrases in our main analyses
(Appendix~\ref{app:detailedbreakdown}, Table~\ref{tab:valsplit}).

\paragraph{Disagreement resolution.} On the 100 overlap sentences per
language, we resolved decisions conservatively: when both validators flagged it with different corrections, we chose between them by
whether the sentence's FLORES+ index was even or odd, a rule fixed before any
corrections were examined
(Appendix~\ref{app:resolution}).

\paragraph{Validation outcomes.} Clean-accept rates were 96.0\% (English),
90.7\% (Russian), and 70.7\% (Serbian); corrections were minor (median 1--3
words). Inter-rater agreement was perfect for English, high for Russian
(Gwet's AC1 \citep{gwet2008computing} 0.87), and moderate for Serbian (AC1 0.63). Full statistics
appear in Appendix~\ref{app:valdetail}.

\subsection{Models and SAEs}
\label{sec:models}

\paragraph{Models.} Our primary model is Gemma-3-27B \citep{gemma3team2025}. We deliberately use standard, publicly released models, SAEs, and labels, since
our aim is to audit the interpretability stack practitioners actually use.
To check that our findings generalize across model scale, we replicate the
core analyses on Gemma-3-1B and Gemma-3-12B; the factorial decomposition
holds across all three sizes (Appendix~\ref{app:scale}). To check generalization across model family we run
Llama-3.1-8B with Llama Scope SAEs \citep{he2024llamascope}, a different model
architecture, SAE training regime, and dictionary width; the qualitative
structure replicates (Appendix~\ref{app:llama}).

 \paragraph{Sparse autoencoders.} We use the Gemma Scope 2 suite of SAEs \citep{gemmascope2_2025}. All SAEs use the JumpReLU activation \citep{rajamanoharan2024}, a dictionary width of $16{,}384$ features, and are selected at low $L_0$ sparsity ($L_0 \in [10, 20]$). We analyze every layer of each model, using the layer-matched SAE at each depth. To confirm our findings are not artifacts of the extraction setup, we vary
dictionary width (Appendix~\ref{app:width}) and $L_0$ sparsity
(Appendix~\ref{app:sparsity}) on Gemma-3-27B and find the decomposition
robust to both. We also compare pooling strategies
(Appendix~\ref{app:pooling}): the ordering holds under last-token and mean
pooling, but only last-token pooling cleanly separates conditions from their
baselines, so we use it throughout.

\subsection{Feature Extraction and Similarity}
\label{sec:features}

\paragraph{Active feature sets.} Given a text $s$, we run it through the model
and, at each layer $\ell$, take the residual-stream hidden state at the final
token position. We pass this hidden state through the layer-$\ell$ SAE encoder,
yielding a feature activation $a_i^{(\ell)}(s)$ for each feature $i$, and define
the active feature set as those that fire:
\[
F_\ell(s) = \{\, i : a_i^{(\ell)}(s) > 0 \,\}.
\]

Under the JumpReLU activation, a feature is nonzero only when its pre-activation exceeds a learned per-feature threshold, so this binary criterion tends to select features that clear a nontrivial gate; we quantify this in Appendix~\ref{app:magnitude}.

\paragraph{Similarity.} We measure the overlap between two texts' active feature sets with the Jaccard index,
\[
J_\ell(s, s') = \frac{|F_\ell(s) \cap F_\ell(s')|}{|F_\ell(s) \cup F_\ell(s')|},
\]
computed independently at each layer.

\paragraph{Factorial comparisons.} We isolate script, language, wording, and
meaning one at a time by comparing pairs that hold the remaining properties
fixed (Table~\ref{tab:factorial}):
\begin{itemize}
\item \textbf{Script:} Sr-Cyrillic orig vs.\ Sr-Latin orig (same language and
content, only script varies).
\item \textbf{Language:} Sr-Cyrillic orig vs.\ Ru-Cyrillic orig (same script
and content, language varies).
\item \textbf{Wording:} English orig vs.\ English paraphrase (same language
and script, surface form varies).
\item \textbf{Meaning:} English orig vs.\ Russian paraphrase (script,
language, and wording all differ; only meaning shared).
\end{itemize}
Each is read against a random-partner baseline sharing only surface
properties, except wording, which uses the identity ceiling of $1.0$. 

\begin{table*}[t]
\centering
\small
\renewcommand{\arraystretch}{1.0}
\setlength{\tabcolsep}{5pt}
\newcommand{\shared}{\ding{51}}
\newcommand{\diff}{\ding{55}}
\begin{tabular}{@{}llcccc@{}}
\toprule
\textbf{Test} & \textbf{Comparison} & \textbf{Script} & \textbf{Language} & \textbf{Wording} & \textbf{Meaning} \\
\midrule
\multirow{2}{*}{\textbf{Script}}
 & Main: Sr-Cyr Orig vs.\ Sr-Lat Orig        & \diff   & \shared & \shared & \shared \\
 & Floor: Sr-Cyr Orig vs.\ Rus-Cyr Rand      & \shared & \diff   & \diff   & \diff   \\
\cmidrule(lr){1-6}
\multirow{2}{*}{\textbf{Language}}
 & Main: Sr-Cyr Orig vs.\ Ru-Cyr Orig        & \shared & \diff   & \shared & \shared \\
 & Floor: Sr-Cyr Orig vs.\ Sr-Lat Rand       & \diff   & \shared & \diff   & \diff   \\
\cmidrule(lr){1-6}
\multirow{2}{*}{\textbf{Wording}}
 & Main: Eng-Lat Orig vs.\ Eng-Lat Para      & \shared & \shared & \diff   & \shared \\
 & Floor: Identity Ceiling ($J{=}1.0$)        & \shared & \shared & \shared & \shared \\
\cmidrule(lr){1-6}
\multirow{2}{*}{\textbf{Meaning}}
 & Main: Eng-Lat Orig vs.\ Rus-Cyr Para      & \diff   & \diff   & \diff   & \shared \\
 & Floor: Eng-Lat Orig vs.\ Rus-Cyr Rand     & \diff   & \diff   & \diff   & \diff   \\
\bottomrule
\end{tabular}
\caption{Controlled comparisons isolating each property of a text.
\ding{51}{} = property \emph{shared} between the two texts and \ding{55}{} =
property \emph{differs}.  Each test isolates one property in its \emph{main} comparison and reads it against
a reference \emph{floor}. Note that the wording comparison uses the identity ceiling rather than a random floor
because two texts cannot differ in meaning while sharing the same wording.}
\label{tab:factorial}
\end{table*}

\subsection{Auto-interpretation evaluation}
\label{sec:methods-detailedbreakdown}

After describing what features collectively encode, we now ask whether the
natural-language labels assigned to individual features hold up. These labels are generated automatically, as described in Section~\ref{sec:intro}, and are often what practitioners rely on to understand and act on a model's internals. A content label makes an implicit prediction: the feature should respond to that
content however it is written. Our paradigm lets us check this directly by
varying surface form while holding meaning fixed. 

\paragraph{Selecting features to test.}
A feature candidate must pass two independent content checks. First, it fires
on both the English original and the Russian paraphrase of a sentence---a
pair that differs in script, language, and wording simultaneously, so
co-activation is strong evidence the feature tracks meaning rather than any
surface property. We anchor on English and Russian rather than either Serbian script, which would
give that script a selection advantage; this leaves both Serbian forms
symmetrically un-anchored. Since the within-language floors are rewordings
of the anchors, absolute cross-language magnitudes should be read as upper
bounds; the Cyrillic/Latin ratio is unaffected. For each such feature, we retrieve its auto-interp label from Neuronpedia, a deployed pipeline serving Gemma Scope~2 descriptions that practitioners consume  \citep{neuronpedia}. We use two labelers, Claude Sonnet (the strongest available) and Gemini Flash Lite (Neuronpedia's default), to confirm the results are not an artifact of one labeler. Labels are retrieved
at the four layers Neuronpedia provides (16, 31, 40, 53), spanning
early-middle to late depth.
From the features that fire on both the English original and the Russian
paraphrase, we apply a second check: a separate classifier LLM (\texttt{claude-sonnet-4-6}, distinct from the auto-interp labelers) classifies each feature's auto-interp label as a \emph{content} claim (semantic topic or meaning) versus a \emph{surface} or \emph{language} claim, and the features whose labels describe surface form rather than content are discarded. This filter is not load-bearing: repeating the analysis on the unfiltered pool yields statistically indistinguishable results, with overlapping confidence intervals at every layer (Appendix~\ref{app:detailedbreakdown}, Table~\ref{tab:27b_ratio_featureCI}). Full classifier details, prompt, and per-layer discard rates are in Appendix~\ref{app:classifier}.

\paragraph{Cross-language falsification.}
For each (feature, sentence) unit we test whether the feature also fires on
renderings of the same sentence in other forms: an English paraphrase and the
Russian original (within-language floors), and the Serbian Latin and Serbian
Cyrillic versions (cross-language test). A content label predicts firing on
all of these; failure on any is a miss. We quantify this per condition: for each rendering $c$, the miss rate is the fraction of
(feature, sentence) units $(i,s)$ in the pool $\mathcal{P}_\ell$ where the
feature fails to fire on that rendering $s_c$:
\begin{equation}
  \mathrm{MR}_\ell(c) = \frac{1}{|\mathcal{P}_\ell|}
  \sum_{(i,s)\in\mathcal{P}_\ell} \mathbf{1}\left[\, i \notin F_\ell(s_c) \,\right]
\end{equation} 
We report the miss rate per condition with 95\% feature-clustered bootstrap confidence intervals (10,000 resamples). The Cyrillic/Latin ratio is bootstrapped within the same feature draw. Flat pair-level resampling gives the same conclusions with narrower intervals; we report the more conservative clustered version. The two
Serbian scripts are deterministic transliterations of each other, so their
miss rates are directly comparable and their ratio isolates script-specific
bias. We report Claude Sonnet as the primary labeler; Gemini gives the same
pattern (Appendix~\ref{app:detailedbreakdown}).

\section{Results}
\label{sec:results}

\subsection{What SAE features encode}
\label{sec:results-leg1}

\begin{figure*}[t]
  \centering
  \includegraphics[width=\textwidth]{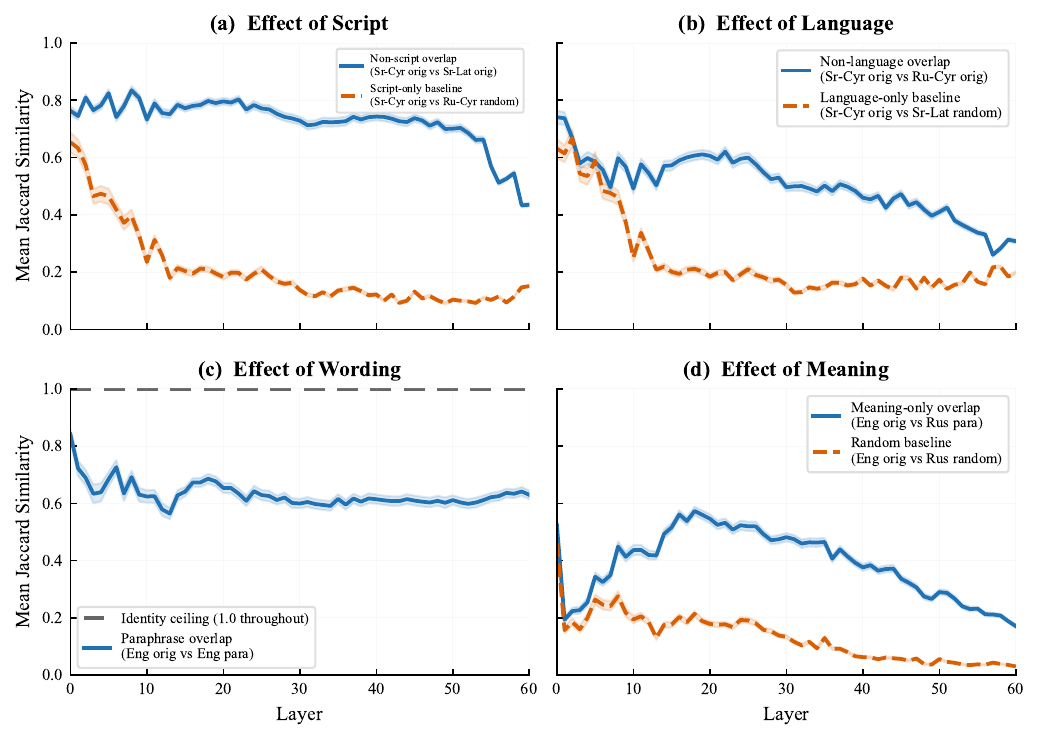}
  \caption{SAE feature overlap decomposed by script, language, wording, and meaning
  across all layers of Gemma-3-27B (300 sentences; shaded bands are 95\% bootstrap
  CIs, 10{,}000 resamples). In each panel a solid \emph{main line} varies only the
  target property while holding the others fixed, read against a dashed
  \emph{baseline} (or, in (c), the identity ceiling); see Table~\ref{tab:factorial}
  for the exact comparisons.}
  \label{fig:decomposition}
\end{figure*}

We first map how a text's active feature set is organized, varying script,
language, wording, and meaning one at a time while holding the others fixed
(Table~\ref{tab:factorial}).
For each property we measure the Jaccard overlap of active SAE feature sets across all
layers of Gemma-3-27B, comparing each against an appropriate reference baseline. We treat script, language, and wording as controls that isolate meaning, the property
central to the rest of our analysis, and summarize these controls briefly before turning to meaning. Figure~\ref{fig:decomposition} shows the full
layer-wise picture: each panel isolates one property, plotting its Jaccard
overlap across all 62 layers.

\paragraph{Script, language, and wording.}
Within our digraphic setting, script has strikingly little effect: transliterating a
sentence between Serbian Latin and Cyrillic holds language, wording, and meaning fixed and
preserves most of the active set (mean Jaccard $0.73$), far above the baseline. In our controlled setting it confirms that
script is the mildest of the four perturbations. Notably, this overlap is negligibly correlated  with the Cyrillic-vs-Latin token-count difference, ruling out tokenization as its driver (Appendix~\ref{app:lentok}).  Language matters more: the same sentence in two languages (Serbian vs.\ Russian, same
script) shares less of its feature set (mean $0.50$), indicating that language has a
larger effect on the active set than script does. Wording sits between the two: paraphrasing within a language preserves a substantial share of the feature
set (mean $0.63$). Each property also changes differently across depth: non-script overlap stays high
until a sharp drop near the output, non-language overlap stays high in the early
layers then declines through the second half, and wording overlap stays roughly flat. The late declines in non-script and non-language overlap are consistent with prior work suggesting that later transformer layers become increasingly specialized for next-token prediction and output generation \citep{geva2022, belrose2023}. We make no strong claim here, however: the corresponding script and language baselines are themselves roughly flat across the later layers, so the picture is suggestive rather than conclusive. These trajectories are worth a detailed study, which the
paradigm supports and we leave to future work. These three properties show how the active set responds to surface variation. We now turn
to the property at the center of our analysis: meaning.

\paragraph{Meaning.}
We isolate meaning by comparing two texts that share nothing but their content---an
English sentence and a Russian paraphrase, differing in script, language, and
wording---so any overlap in their feature sets points to genuinely cross-lingual
semantic features rather than surface form. Even with script, language, and wording all changed at once, overlap holds at a 0.39 Jaccard average, well above the baseline (mean 0.13), and peaks at 0.57 in the lower-middle layers
(Figure~\ref{fig:decomposition}d). SAE features therefore appear to encode abstract meaning that largely survives a complete change of surface form. The signal is also depth-dependent: after an initial drop in the first layer or two, the
overlap climbs to a peak around the lower-middle of the network before declining through the later layers.

\paragraph{From feature sets to feature labels.}
Having established that genuinely semantic features exist, we next need a way to identify
what each one represents. Across thousands of features, this is typically done with auto-interpretation, which
assigns each feature a natural-language label. Our set-level analysis does not address this: it describes populations of features, not whether any individual feature's label holds. We turn to that question next.

\subsection{Auto-interpretation labels generalize poorly across languages and scripts}
\label{sec:results-leg2}
Leg~1 gave a set-level picture: SAE features appear to encode genuine semantic content, while
script, wording, and  language reshape the active set to differing degrees. Auto-interpretation provides a feature-level view, labeling each feature from its
top-activating examples. We evaluate these labels on the cross-lingual semantic features
identified in Leg~1: those that fire on a sentence's content in both English and Russian,
which confirms behaviorally that they track meaning rather than surface form. As a second robustness check, we confirm these features' labels describe content rather
than surface form, using an LLM to classify each label as content or surface and
discarding the  exceptions (Appendices~\ref{app:classifier} and~\ref{app:detailedbreakdown}).

Testing labels from Neuronpedia's auto-interpretation pipeline, where Claude Sonnet and Gemini Flash Lite serve as the labelers, we find these predictions often break down: a feature labeled for its content can fail
to fire on that same content when it is rewritten in another language or script, and
this happens more for less well-represented forms. We quantify this as the \emph{miss rate}: the fraction of cases where a content-labeled
feature fails to fire on the same meaning in another form (Table~\ref{tab:leg2}). We
report Claude Sonnet labels, the stronger labeler; Gemini gives the same picture,
converging on near-identical Cyrillic/Latin ratios at every layer (e.g.,
1.35$\times$ vs.\ 1.36$\times$ at layer 53), which suggests the failure lies in the
features rather than in either labeler. The full breakdown and additional baselines
(including random controls) are in Appendix~\ref{app:detailedbreakdown}.

\begin{table}[t]
\centering
\small
\setlength{\tabcolsep}{2pt}

\begin{subtable}[t]{\columnwidth}
\centering
\begin{tabular}{@{}cccc@{}}
\toprule
Layer & Eng-para (\%) & Rus-orig (\%) & Serbian avg (\%) \\
\midrule
16 & 14.8 [12.4, 18.6] & 16.6 [12.5, 22.0] & 21.0 [15.8, 28.2] \\
31 &  9.2 [7.0, 11.7]  & 13.7 [11.4, 16.6] & 21.5 [19.4, 24.1] \\
40 &  9.1 [7.7, 10.5]  & 15.6 [14.0, 17.3] & 24.1 [21.8, 26.5] \\
53 &  7.3 [5.8, 9.0]   & 15.4 [13.3, 17.6] & 30.1 [26.6, 33.8] \\
\bottomrule
\end{tabular}
\caption{Within- vs.\ cross-language miss rates.}
\label{tab:leg2a}
\end{subtable}

\vspace{0.8em}

\begin{subtable}[t]{\columnwidth}
\centering
\begin{tabular}{@{}cccc@{}}
\toprule
Layer & Sr-Lat (\%) & Sr-Cyr (\%) & Cyr/Lat \\
\midrule
16 & 20.8 [15.8, 27.5] & 21.3 [15.7, 29.0] & 1.02$\times$ [0.91, 1.12] \\
31 & 20.4 [18.2, 23.4] & 22.6 [20.4, 25.1] & 1.11$\times$ [1.01, 1.20] \\
40 & 22.2 [19.9, 24.6] & 25.9 [23.4, 28.7] & 1.17$\times$ [1.09, 1.26] \\
53 & 25.6 [22.1, 29.6] & 34.6 [30.7, 38.6] & 1.35$\times$ [1.23, 1.48] \\
\bottomrule
\end{tabular}
\caption{Serbian script asymmetry.}
\label{tab:leg2b}
\end{subtable}

\caption{Content-feature miss rates over the cross-lingual content pool. (a) Within-language
floors (Eng-para, Rus-orig) vs.\ the cross-language Serbian average. (b) Serbian Latin vs.\
Cyrillic, with the Cyrillic/Latin ratio. 95\% bootstrap CIs in brackets ($10{,}000$ resamples, feature-clustered).}
\label{tab:leg2}
\end{table}

\paragraph{Even within a language, content labels are unreliable; across languages they break down.}
Reworded within their own language, these features already miss a non-trivial share of
the time, from 14.8\% in early layers to 7.3\% at layer 53 for
English, and staying in the 13--17\% range for Russian. The same features miss
Serbian far more often. Averaged over the two Serbian scripts, miss rates start near the
within-language floor (21.0\% versus 14.8\% at layer~16) but reach 30.1\% by layer~53 (Table~\ref{tab:leg2a})---an excess of more than 20 points over the
English floor, roughly 4$\times$ the within-language miss rate. Paraphrasing the Serbian text differs only by 1 to 3 points
(Appendix~\ref{app:detailedbreakdown}), confirming the failure comes from the language and
script, not the specific wording within them. Across conditions, miss rates are ordered from English through Russian and Serbian Latin to Serbian Cyrillic (Table~\ref{tab:leg2}).

\paragraph{The Serbian Cyrillic disadvantage grows with depth.}
The two Serbian scripts are deterministic transliterations of each other, so a content
feature should treat them identically. Instead they diverge with depth: indistinguishable
in early layers (1.02$\times$ Cyrillic-to-Latin miss ratio at layer~16), the gap rises
to 1.35$\times$ at layer~53 (Table~\ref{tab:leg2b}). Since the scripts share language, wording, and meaning, the failure is unlikely to be
semantic; the main difference is the character system, which points to how well the model
represents each script. The same gap appears whether we test the original Serbian sentences or their
paraphrases, so it does not depend on the particular wording. The Cyrillic disadvantage holds within every topic category (Appendix~\ref{app:detailedbreakdown}), confirming it is not specific to any particular domain. Table~\ref{tab:vignettes} gives individual examples: content-labeled features that fire
on a sentence in three of the four forms but fail on one Serbian script. A manual check of a sample of missed features confirms that the labels are
largely accurate on their own top-activating examples
(Appendix~\ref{app:humanverif}).

\begin{table}[t]
\centering
\small
\begin{tabular}{@{}p{\columnwidth}@{}}
\toprule
\textbf{L40 / F4327} --- misses Sr-Cyrillic \\
\quad \textbf{Auto-interp:} Sonnet: ``Olympic athletes'' | Gemini: ``Olympic athletes and comebacks'' \\
\quad \textbf{Sentence:} ``Poland's visually impaired skier Maciej Krezel\ldots'' \\
\midrule
\textbf{L53 / F3041} --- misses Sr-Cyrillic \\
\quad \textbf{Auto-interp:} Sonnet: ``political leaders'' | Gemini: ``political leaders or figures'' \\
\quad \textbf{Sentence:} ``Hu encouraged developing countries to avoid the old path\ldots'' \\
\midrule
\textbf{L40 / F8424} --- misses Sr-Latin \\
\quad \textbf{Auto-interp:} Sonnet: ``diseases and conditions'' | Gemini: ``diseases and conditions'' \\
\quad \textbf{Sentence:} ``Infectious diseases themselves, or dangerous animals\ldots'' \\
\bottomrule
\end{tabular}
\caption{Content-labeled features that fail on one Serbian script, verifiable on
Neuronpedia. Additional examples in Appendix~\ref{app:detailedbreakdown}, Table~\ref{tab:vignettes-appendix}.
\label{tab:vignettes}}
\end{table}

\paragraph{The divergence sharpens with depth.}
These failures are mildest in the earliest layer we can probe: at layer~16 the
within-language floors and the Serbian miss rate sit fairly close ($14$--$17\%$ versus
$21.0\%$). They pull apart deeper in the network. The within-language floors hold steady or fall while the Serbian miss rate
climbs (Table~\ref{tab:leg2}a). Reliability thus holds or
improves within a language and degrades across one as depth increases. These patterns replicate across model scale: Gemma-3-1B and Gemma-3-12B show
the same ordering and a Cyrillic/Latin asymmetry that grows with depth
(Appendix~\ref{app:leg2_scale}).

\section{Discussion}
\label{sec:discussion}

\paragraph{Semantic features are real, but their labels overpromise.}
Our two analyses give a coherent picture. Leg~1 indicates that SAE features genuinely encode
abstract meaning: cross-lingual semantic overlap exists and survives a complete change of
script, language, and wording. Leg~2 shows that this encoding is narrower than its labels
suggest. A feature that tracks a concept in both English and Russian, and is labeled for
that concept, still fails to fire on it in Serbian noticeably more often than it does
within those languages. The two findings are not in tension: these features carry real
semantic content, which is why their labels can be reasonable, but that content does
not transfer cleanly across languages and scripts. The failure is not that the labels are always wrong: they often capture something real about
the feature, yet can still fail to predict its behavior on an equivalent rendering of the
same content, with no indication of where.

\paragraph{The failures are consistent with expected training distribution.}
The pattern of failures is consistent with how often each form of a meaning is likely to appear in pretraining.  Averaged
across layers, content features miss their target far more in Serbian than within English
or Russian (10.1\% and 15.3\% versus 22.2\% for Serbian Latin and 26.1\% for
Serbian Cyrillic). The clearest case is the Serbian script asymmetry: the two scripts are
deterministic transliterations, identical in language, wording, and meaning, so a
difference between them cannot be semantic. The only thing that differs is the character
system, and features miss Serbian Cyrillic more than Serbian Latin. This is consistent with broad patterns in web text. Web corpora are heavily English-dominant: English is the primary language of roughly 40--45\% of HTML pages in recent Common Crawl snapshots, and no other language exceeds about 7\% \citep{commoncrawl}. Russian, the second-largest, accounts for roughly 6--7\% of pages versus about 0.2--0.27\% for Serbian \citep{commoncrawl}, while Latin script is estimated to dominate Serbian usage (roughly 70:30 Latin to Cyrillic) \citep{serbian_domains}. Together, these factors suggest that Serbian Cyrillic is comparatively scarce in the web text that dominates modern pretraining corpora. Such imbalances have measurable downstream effects: model performance has been shown to
align with a language's share of pretraining data \citep{li2024}. A similar pattern could hold at the level of individual features: the Serbian Cyrillic
gap is what this predicts, with the rarest form of the content the one its features
handle worst. This is not a binary high- versus
low-resource split but a graded one---English is the most dominant form, Russian
high-resource but less dominant, and Serbian rarer still---with reliability ordered
accordingly.

\paragraph{How the gap changes with depth.}
As Section~\ref{sec:results-leg2} shows, this ordering sharpens with depth. One reading is that deeper layers grow less language-agnostic, re-specializing toward specific surface forms \citep{wendler2024, liu2025middlelayer}; if that specialization favors the forms a model has seen most, content features could fire more reliably on well-represented inputs and less on rarer ones, as we observe; however, we cannot establish this directly. The within-language trend, however, is notable: a feature becoming \emph{more} reliable for its strongest form with depth is, to our knowledge, new, and we offer it as an empirical observation. One speculative possibility is that this is a depth effect: where early layers respond more uniformly across a meaning's different language and script renderings, deeper layers concentrate firing on the renderings the model has seen most; we leave testing this to future work.

None of this is visible from the label itself. Consider a feature labeled ``deception'' or ``violent content.'' A researcher
auditing the model reads the label, takes the feature to track that concept, and relies
on it---perhaps to monitor or steer the behavior. What the label would not reveal is that such a feature might detect the concept mainly in
English, with its response weakening in Russian and weakening further in Serbian Cyrillic,
so the same content that triggers it in one form could pass unnoticed in another. The label may name the concept correctly; what it can omit is the range over which the
feature actually tracks it.

\paragraph{Implications for practitioners.}
The practical lesson is that an auto-interpretation label guarantees nothing about how a feature behaves on the same concept in another form. A content label can be locally accurate and still fail to generalize
across languages and scripts, and because the label carries no signal of
this, the failure is invisible to monolingual inspection. Evaluating feature
labels therefore requires controlled stimuli that vary surface form while
holding meaning fixed. The failures we document also appear addressable:
auto-interpretation pipelines could draw top-activating examples from multiple
languages, or flag features whose activation is inconsistent across equivalent
inputs, though we leave validation of such mitigations to future work.

\paragraph{Concurrent and recent work.}
The interpretability toolchain is developing rapidly. Circuit tracing
\citep{ameisen2025circuit} traces computational graphs through model
internals, and Natural Language Autoencoders \citep{fraser-taliente2026nla} translate full
activation vectors into natural-language explanations. These efforts build increasingly powerful tools for inspecting
models; our work is complementary, contributing controlled evaluation
methodology that checks whether the descriptions these tools produce
generalize beyond the conditions they were derived from. Both NLAs and the
auto-interp labels we evaluate face the same underlying challenge: a
natural-language description of a model's internals may be accurate on the
inputs it was built from and quietly fail on others.

\section{Conclusion}

Using Serbian digraphia as a controlled testbed, we find evidence that SAE
features in multilingual language models encode abstract meaning that can
survive complete changes of script, language, and wording. However, the
auto-interpretation labels assigned to these features do not keep pace: content
labels that hold in well-represented languages quietly fail on
less-represented ones, with miss rates reaching $4\times$ the within-language
floor for Serbian, and the labels themselves carry no signal of where
they break. Even between Serbian Cyrillic and Latin, identical up to transliteration, miss rates diverge by up to 1.35$\times$, a gap that cannot be semantic. The failures are graded, aligning with the estimated representation of each form in
training data, and tend to sharpen with network depth. These results suggest that
auto-interpretation labels should be understood as claims about a feature's
behavior on its best-represented inputs, not as guarantees about the concept
in general.

\section*{Limitations}

Our cross-lingual evaluation covers three languages (English, Russian, and
Serbian) from two language families, with Serbian closely related to Russian.
The graded miss-rate pattern we observe should in principle extend to more
typologically distant and lower-resource languages, but we have not verified
this beyond the conditions in our dataset. 

Our sample of 300 sentences is small, a trade-off we made to prioritize
validation quality; scaling it with LLM-generated data is a natural extension, for which our human-validated set can serve as a reference to measure the reliability of fully automated generation. 

We also treat each missed feature in isolation: when a specific feature fails to
fire, a more general related feature may fire in its place, possibly reflecting
lower confidence on less-represented inputs rather than lost content. Examining
which features fire instead is a natural direction for future work. 

Leg~2 replicates across Gemma
model scales (1B, 12B, 27B; Appendix~\ref{app:leg2_scale}), but we have not
confirmed the label-failure patterns on other model families or labeling
pipelines beyond Neuronpedia. Within Neuronpedia, we also hold the labeling procedure fixed: two independent
labelers agree, but we do not vary how many top-activating examples each sees,
which we leave to future work.

\section*{Ethics Statement}

Native-speaker validators were recruited through personal networks and
compensated \$125--\$200 per batch of 200 sentence pairs, exceeding standard
crowdsourcing rates for comparable linguistic annotation. No personal or
sensitive data was collected from participants. Paraphrases were generated
using Claude Opus 4.6 and auto-interpretation labels were retrieved from
Neuronpedia's deployed pipeline (Claude Sonnet and Gemini Flash Lite); all prompts
are reported in full in Appendices~\ref{app:prompts} and~\ref{app:classifier}. The FLORES+ source data is used under its
CC BY-SA 4.0 license. Our released dataset and code are available under CC BY
4.0 for research use.

\paragraph{Use of Large Language Models.} Claude was used to assist with
paper drafting, editing, and code implementation. All experimental design,
analysis, and scientific claims are the author's own.

\section*{Acknowledgments}
I thank Rejn Omeragic and Nazgul Maksutkhan for helping me recruit native
speakers of Russian and Serbian, and I thank the validators who made the
full multilingual evaluation possible: Samuel Dudkin and Lennard Teslia (Russian),
Elena and Jovana (Serbian), and Tharangini Jakkena, Ram Karne, and Sai Srikar
Karne (English). I also thank Tianyi Lorena Yan and Jonathn Chang for reviewing
the final draft and providing valuable feedback.

\bibliography{custom}

\begin{thebibliography}{32}
\providecommand{\natexlab}[1]{#1}

\bibitem[{Ameisen et~al.(2025)Ameisen, Lindsey, Pearce, Gurnee, Turner, Chen, Citro, Abrahams, Carter, Hosmer, Marcus, Sklar, Templeton, Bricken, McDougall, Cunningham, Henighan, Jermyn, Jones, Persic, Qi, Thompson, Zimmerman, Rivoire, Conerly, Olah, and Batson}]{ameisen2025circuit}
Emmanuel Ameisen, Jack Lindsey, Adam Pearce, Wes Gurnee, Nicholas~L. Turner, Brian Chen, Craig Citro, David Abrahams, Shan Carter, Basil Hosmer, Jonathan Marcus, Michael Sklar, Adly Templeton, Trenton Bricken, Callum McDougall, Hoagy Cunningham, Thomas Henighan, Adam Jermyn, Andy Jones, and 8 others. 2025.
\newblock \href {https://transformer-circuits.pub/2025/attribution-graphs/methods.html} {Circuit tracing: Revealing computational graphs in language models}.
\newblock \emph{Transformer Circuits Thread}.

\bibitem[{Belrose et~al.(2023)Belrose, Ostrovsky, McKinney, Furman, Smith, Halawi, Biderman, and Steinhardt}]{belrose2023}
Nora Belrose, Igor Ostrovsky, Lev McKinney, Zach Furman, Logan Smith, Danny Halawi, Stella Biderman, and Jacob Steinhardt. 2023.
\newblock \href {https://arxiv.org/abs/2303.08112} {Eliciting latent predictions from transformers with the tuned lens}.
\newblock \emph{arXiv preprint arXiv:2303.08112}.

\bibitem[{Bills et~al.(2023)Bills, Cammarata, Mossing, Tillman, Gao, Goh, Sutskever, Leike, Wu, and Saunders}]{bills2023language}
Steven Bills, Nick Cammarata, Dan Mossing, Henk Tillman, Leo Gao, Gabriel Goh, Ilya Sutskever, Jan Leike, Jeff Wu, and William Saunders. 2023.
\newblock Language models can explain neurons in language models.
\newblock \url{https://openaipublic.blob.core.windows.net/neuron-explainer/paper/index.html}.

\bibitem[{Bricken et~al.(2023)Bricken, Templeton, Batson, Chen, Jermyn, Conerly, Turner, Anil, Denison, Askell, Lasenby, Wu, Kravec, Schiefer, Maxwell, Joseph, Tamkin, Nguyen, McLean, Burke, Hume, Carter, Henighan, and Olah}]{bricken2023}
Trenton Bricken, Adly Templeton, Joshua Batson, Brian Chen, Adam Jermyn, Tom Conerly, Nicholas~L. Turner, Cem Anil, Carson Denison, Amanda Askell, Robert Lasenby, Yifan Wu, Shauna Kravec, Nicholas Schiefer, Tim Maxwell, Nicholas Joseph, Alex Tamkin, Karina Nguyen, Brayden McLean, and 5 others. 2023.
\newblock \href {https://transformer-circuits.pub/2023/monosemantic-features/index.html} {Towards monosemanticity: Decomposing language models with dictionary learning}.
\newblock \emph{Transformer Circuits Thread}.

\bibitem[{Brinkmann et~al.(2025)Brinkmann, Wendler, Bartelt, and Mueller}]{brinkmann2025large}
Jannik Brinkmann, Chris Wendler, Christian Bartelt, and Aaron Mueller. 2025.
\newblock \href {https://doi.org/10.18653/v1/2025.naacl-long.312} {Large language models share representations of latent grammatical concepts across typologically diverse languages}.
\newblock In \emph{Proceedings of the 2025 Conference of the Nations of the Americas Chapter of the Association for Computational Linguistics: Human Language Technologies (Volume 1: Long Papers)}, pages 6131--6150, Albuquerque, New Mexico. Association for Computational Linguistics.

\bibitem[{Burchell et~al.(2024)Burchell, Maillard, Anastasopoulos, Federmann, Koehn, and Wang}]{maillard-etal-2024-findings}
Laurie Burchell, Jean Maillard, Antonios Anastasopoulos, Christian Federmann, Philipp Koehn, and Skyler Wang. 2024.
\newblock \href {https://doi.org/10.18653/v1/2024.wmt-1.4} {Findings of the {WMT} 2024 shared task of the open language data initiative}.
\newblock In \emph{Proceedings of the Ninth Conference on Machine Translation}, pages 110--117, Miami, Florida, USA. Association for Computational Linguistics.

\bibitem[{{Common Crawl Foundation}(2026)}]{commoncrawl}
{Common Crawl Foundation}. 2026.
\newblock \href {https://commoncrawl.github.io/cc-crawl-statistics/plots/languages.html} {Statistics of {Common Crawl} monthly archives: Distribution of languages}.
\newblock Accessed 2026-09-30.

\bibitem[{Cunningham et~al.(2024)Cunningham, Ewart, Riggs, Huben, and Sharkey}]{cunningham2024}
Hoagy Cunningham, Aidan Ewart, Logan Riggs, Robert Huben, and Lee Sharkey. 2024.
\newblock \href {https://proceedings.iclr.cc/paper_files/paper/2024/hash/1fa1ab11f4bd5f94b2ec20e794dbfa3b-Abstract-Conference.html} {Sparse autoencoders find highly interpretable features in language models}.
\newblock In \emph{The Twelfth International Conference on Learning Representations}.

\bibitem[{Deng et~al.(2025)Deng, Wan, Yang, Zhang, and Feng}]{deng2025unveiling}
Boyi Deng, Yu~Wan, Baosong Yang, Yidan Zhang, and Fuli Feng. 2025.
\newblock \href {https://doi.org/10.18653/v1/2025.acl-long.229} {Unveiling language-specific features in large language models via sparse autoencoders}.
\newblock In \emph{Proceedings of the 63rd Annual Meeting of the Association for Computational Linguistics (Volume 1: Long Papers)}, pages 4563--4608, Vienna, Austria. Association for Computational Linguistics.

\bibitem[{Feng et~al.(2022)Feng, Yang, Cer, Arivazhagan, and Wang}]{feng-etal-2022-language}
Fangxiaoyu Feng, Yinfei Yang, Daniel Cer, Naveen Arivazhagan, and Wei Wang. 2022.
\newblock \href {https://doi.org/10.18653/v1/2022.acl-long.62} {Language-agnostic {BERT} sentence embedding}.
\newblock In \emph{Proceedings of the 60th Annual Meeting of the Association for Computational Linguistics (Volume 1: Long Papers)}, pages 878--891, Dublin, Ireland. Association for Computational Linguistics.

\bibitem[{Fraser-Taliente et~al.(2026)Fraser-Taliente, Kantamneni, Ong, Mossing, Lu, Bogdan, Ameisen, Chen, Kishylau, Pearce, Tarng, Wu, Wu, Zhang, Ziegler, Hubinger, Batson, Lindsey, Zimmerman, and Marks}]{fraser-taliente2026nla}
Kit Fraser-Taliente, Subhash Kantamneni, Euan Ong, Dan Mossing, Christina Lu, Paul~C. Bogdan, Emmanuel Ameisen, James Chen, Dzmitry Kishylau, Adam Pearce, Julius Tarng, Alex Wu, Jeff Wu, Yang Zhang, Daniel~M. Ziegler, Evan Hubinger, Joshua Batson, Jack Lindsey, Samuel Zimmerman, and Samuel Marks. 2026.
\newblock \href {https://transformer-circuits.pub/2026/nla/index.html} {Natural language autoencoders produce unsupervised explanations of {LLM} activations}.
\newblock \emph{Transformer Circuits Thread}.

\bibitem[{Gao et~al.(2025)Gao, {Dupr{\'e} la Tour}, Tillman, Goh, Troll, Radford, Sutskever, Leike, and Wu}]{gao2024scaling}
Leo Gao, Tom {Dupr{\'e} la Tour}, Henk Tillman, Gabriel Goh, Rajan Troll, Alec Radford, Ilya Sutskever, Jan Leike, and Jeffrey Wu. 2025.
\newblock \href {https://openreview.net/forum?id=tcsZt9ZNKD} {Scaling and evaluating sparse autoencoders}.
\newblock In \emph{The Thirteenth International Conference on Learning Representations}.

\bibitem[{{Gemma Team}(2025)}]{gemma3team2025}
{Gemma Team}. 2025.
\newblock \href {https://arxiv.org/abs/2503.19786} {Gemma 3 technical report}.
\newblock \emph{Preprint}, arXiv:2503.19786.

\bibitem[{Geva et~al.(2022)Geva, Caciularu, Wang, and Goldberg}]{geva2022}
Mor Geva, Avi Caciularu, Kevin Wang, and Yoav Goldberg. 2022.
\newblock \href {https://doi.org/10.18653/v1/2022.emnlp-main.3} {Transformer feed-forward layers build predictions by promoting concepts in the vocabulary space}.
\newblock In \emph{Proceedings of the 2022 Conference on Empirical Methods in Natural Language Processing}, pages 30--45, Abu Dhabi, United Arab Emirates. Association for Computational Linguistics.

\bibitem[{Goyal et~al.(2022)Goyal, Gao, Chaudhary, Chen, Wenzek, Ju, Krishnan, Ranzato, Guzm{\'a}n, and Fan}]{goyal-etal-2022-flores}
Naman Goyal, Cynthia Gao, Vishrav Chaudhary, Peng-Jen Chen, Guillaume Wenzek, Da~Ju, Sanjana Krishnan, Marc{'}Aurelio Ranzato, Francisco Guzm{\'a}n, and Angela Fan. 2022.
\newblock \href {https://doi.org/10.1162/tacl_a_00474} {The {F}lores-101 evaluation benchmark for low-resource and multilingual machine translation}.
\newblock \emph{Transactions of the Association for Computational Linguistics}, 10:522--538.

\bibitem[{Gwet(2008)}]{gwet2008computing}
Kilem~Li Gwet. 2008.
\newblock \href {https://doi.org/10.1348/000711006X126600} {Computing inter-rater reliability and its variance in the presence of high agreement}.
\newblock \emph{British Journal of Mathematical and Statistical Psychology}, 61(1):29--48.

\bibitem[{He et~al.(2024)He, Shu, Ge, Chen, Wang, Zhou, Liu, Guo, Huang, Wu, Jiang, and Qiu}]{he2024llamascope}
Zhengfu He, Wentao Shu, Xuyang Ge, Lingjie Chen, Junxuan Wang, Yunhua Zhou, Frances Liu, Qipeng Guo, Xuanjing Huang, Zuxuan Wu, Yu-Gang Jiang, and Xipeng Qiu. 2024.
\newblock \href {https://arxiv.org/abs/2410.20526} {{Llama Scope}: Extracting millions of features from {Llama-3.1-8B} with sparse autoencoders}.
\newblock \emph{arXiv preprint arXiv:2410.20526}.

\bibitem[{Huang et~al.(2023)Huang, Geiger, D{'}Oosterlinck, Wu, and Potts}]{huang2023rigorously}
Jing Huang, Atticus Geiger, Karel D{'}Oosterlinck, Zhengxuan Wu, and Christopher Potts. 2023.
\newblock \href {https://doi.org/10.18653/v1/2023.blackboxnlp-1.24} {Rigorously assessing natural language explanations of neurons}.
\newblock In \emph{Proceedings of the 6th BlackboxNLP Workshop: Analyzing and Interpreting Neural Networks for NLP}, pages 317--331, Singapore. Association for Computational Linguistics.

\bibitem[{Kopf et~al.(2025)Kopf, Feldhus, Bykov, Bommer, Hedstr{\"o}m, H{\"o}hne, and Eberle}]{kopf2025prism}
Laura Kopf, Nils Feldhus, Kirill Bykov, Philine~Lou Bommer, Anna Hedstr{\"o}m, Marina M.-C. H{\"o}hne, and Oliver Eberle. 2025.
\newblock \href {https://doi.org/10.52202/085713-2780} {Capturing polysemanticity with {PRISM}: A multi-concept feature description framework}.
\newblock In \emph{Advances in Neural Information Processing Systems}, volume~38, pages 82938--82974.

\bibitem[{Li et~al.(2025)Li, Shi, Liu, Yang, Payani, Liu, and Du}]{li2024}
Zihao Li, Yucheng Shi, Zirui Liu, Fan Yang, Ali Payani, Ninghao Liu, and Mengnan Du. 2025.
\newblock \href {https://doi.org/10.1609/aaai.v39i27.35038} {Language ranker: A metric for quantifying {LLM} performance across high and low-resource languages}.
\newblock In \emph{Proceedings of the AAAI Conference on Artificial Intelligence}, volume~39, pages 28186--28194.

\bibitem[{Lin(2023)}]{neuronpedia}
Johnny Lin. 2023.
\newblock \href {https://www.neuronpedia.org} {Neuronpedia: Interactive reference and tooling for analyzing neural networks}.
\newblock Software available from neuronpedia.org.

\bibitem[{Liu and Niehues(2025)}]{liu2025middlelayer}
Danni Liu and Jan Niehues. 2025.
\newblock \href {https://doi.org/10.18653/v1/2025.acl-long.778} {Middle-layer representation alignment for cross-lingual transfer in fine-tuned {LLM}s}.
\newblock In \emph{Proceedings of the 63rd Annual Meeting of the Association for Computational Linguistics (Volume 1: Long Papers)}, pages 15979--15996, Vienna, Austria. Association for Computational Linguistics.

\bibitem[{Liu et~al.(2026)Liu, Miao, Zhao, Liu, and Du}]{liu2026neuronscope}
Weiqi Liu, Yongliang Miao, Haiyan Zhao, Yanguang Liu, and Mengnan Du. 2026.
\newblock \href {https://arxiv.org/abs/2601.03671} {{NeuronScope}: A multi-agent framework for explaining polysemantic neurons in language models}.
\newblock \emph{arXiv preprint arXiv:2601.03671}.

\bibitem[{McDougall et~al.(2025)McDougall, Conmy, Kram{\'a}r, Lieberum, Rajamanoharan, and Nanda}]{gemmascope2_2025}
Callum McDougall, Arthur Conmy, J{\'a}nos Kram{\'a}r, Tom Lieberum, Senthooran Rajamanoharan, and Neel Nanda. 2025.
\newblock \href {https://storage.googleapis.com/deepmind-media/DeepMind.com/Blog/gemma-scope-2-helping-the-ai-safety-community-deepen-understanding-of-complex-language-model-behavior/Gemma_Scope_2_Technical_Paper.pdf} {{Gemma Scope 2} - technical paper}.
\newblock Technical report, Google DeepMind.

\bibitem[{{NLLB Team} et~al.(2024){NLLB Team}, Costa-juss{\`a}, Cross, {\c{C}}elebi, Elbayad, Heafield, Heffernan, Kalbassi, Lam, Licht, Maillard, Sun, Wang, Wenzek, Youngblood, Akula, Barrault, Mejia~Gonzalez, Hansanti, Hoffman, Jarrett, Ram~Sadagopan, Rowe, Spruit, Tran, Andrews, Ayan, Bhosale, Edunov, Fan, Gao, Goswami, Guzm{\'a}n, Koehn, Mourachko, Ropers, Saleem, Schwenk, and Wang}]{nllb-2024}
{NLLB Team}, Marta~R. Costa-juss{\`a}, James Cross, Onur {\c{C}}elebi, Maha Elbayad, Kenneth Heafield, Kevin Heffernan, Elahe Kalbassi, Janice Lam, Daniel Licht, Jean Maillard, Anna Sun, Skyler Wang, Guillaume Wenzek, Al~Youngblood, Bapi Akula, Lo{\"i}c Barrault, Gabriel Mejia~Gonzalez, Prangthip Hansanti, and 20 others. 2024.
\newblock \href {https://doi.org/10.1038/s41586-024-07335-x} {Scaling neural machine translation to 200 languages}.
\newblock \emph{Nature}, 630(8018):841--846.

\bibitem[{Paulo et~al.(2025)Paulo, Mallen, Juang, and Belrose}]{paulo2025automatically}
Gon{\c{c}}alo~Santos Paulo, Alex~Troy Mallen, Caden Juang, and Nora Belrose. 2025.
\newblock \href {https://proceedings.mlr.press/v267/paulo25a.html} {Automatically interpreting millions of features in large language models}.
\newblock In \emph{Proceedings of the 42nd International Conference on Machine Learning}, volume 267 of \emph{Proceedings of Machine Learning Research}, pages 48393--48421. PMLR.

\bibitem[{Rajamanoharan et~al.(2024)Rajamanoharan, Lieberum, Sonnerat, Conmy, Varma, Kram{\'a}r, and Nanda}]{rajamanoharan2024}
Senthooran Rajamanoharan, Tom Lieberum, Nicolas Sonnerat, Arthur Conmy, Vikrant Varma, J{\'a}nos Kram{\'a}r, and Neel Nanda. 2024.
\newblock \href {https://arxiv.org/abs/2407.14435} {Jumping ahead: Improving reconstruction fidelity with {JumpReLU} sparse autoencoders}.
\newblock \emph{Preprint}, arXiv:2407.14435.

\bibitem[{{Serbian National Internet Domain Registry (RNIDS)}(2016)}]{serbian_domains}
{Serbian National Internet Domain Registry (RNIDS)}. 2016.
\newblock \href {https://www.rnids.rs/en/news/cyrbusters-panel-debate-held-busting-myths-about-cyrillic-on-the-internet-2} {Cyrbusters panel debate held: Busting myths about {Cyrillic} on the {Internet}}.
\newblock RNIDS news, 28 January 2016.
\newblock Accessed 2026-09-30.

\bibitem[{Templeton et~al.(2024)Templeton, Conerly, Marcus, Lindsey, Bricken, Chen, Pearce, Citro, Ameisen, Jones, Cunningham, Turner, McDougall, MacDiarmid, Tamkin, Durmus, Hume, Mosconi, Freeman, Sumers, Rees, Batson, Jermyn, Carter, Olah, and Henighan}]{templeton2024}
Adly Templeton, Tom Conerly, Jonathan Marcus, Jack Lindsey, Trenton Bricken, Brian Chen, Adam Pearce, Craig Citro, Emmanuel Ameisen, Andy Jones, Hoagy Cunningham, Nicholas~L. Turner, Callum McDougall, Monte MacDiarmid, Alex Tamkin, Esin Durmus, Tristan Hume, Francesco Mosconi, C.~Daniel Freeman, and 7 others. 2024.
\newblock \href {https://transformer-circuits.pub/2024/scaling-monosemanticity/index.html} {Scaling monosemanticity: Extracting interpretable features from {Claude 3 Sonnet}}.
\newblock \emph{Transformer Circuits Thread}.

\bibitem[{Verma et~al.(2026)Verma, Chatterjee, Gupta, and Chakraborty}]{verma2026}
Aastha A~K Verma, Anwoy Chatterjee, Mehak Gupta, and Tanmoy Chakraborty. 2026.
\newblock \href {https://doi.org/10.18653/v1/2026.acl-long.1554} {Multilingual language models encode script over linguistic structure}.
\newblock In \emph{Proceedings of the 64th Annual Meeting of the {A}ssociation for {C}omputational {L}inguistics (Volume 1: Long Papers)}, pages 33685--33719, San Diego, California, United States. Association for Computational Linguistics.

\bibitem[{Wendler et~al.(2024)Wendler, Veselovsky, Monea, and West}]{wendler2024}
Chris Wendler, Veniamin Veselovsky, Giovanni Monea, and Robert West. 2024.
\newblock \href {https://doi.org/10.18653/v1/2024.acl-long.820} {Do llamas work in {E}nglish? on the latent language of multilingual transformers}.
\newblock In \emph{Proceedings of the 62nd Annual Meeting of the Association for Computational Linguistics (Volume 1: Long Papers)}, pages 15366--15394, Bangkok, Thailand. Association for Computational Linguistics.

\bibitem[{Wu et~al.(2025)Wu, Yu, Yogatama, Lu, and Kim}]{wu2025semantichub}
Zhaofeng Wu, Xinyan~Velocity Yu, Dani Yogatama, Jiasen Lu, and Yoon Kim. 2025.
\newblock \href {https://proceedings.iclr.cc/paper_files/paper/2025/hash/84fb08ae30ab266953d1ff72eb0611b1-Abstract-Conference.html} {The semantic hub hypothesis: Language models share semantic representations across languages and modalities}.
\newblock In \emph{The Thirteenth International Conference on Learning Representations}, pages 53705--53723.

\end{thebibliography}

\appendix

\section{Dataset Details}
\label{app:dataset}

\subsection{Paraphrase Generation}
\label{app:prompts}

All paraphrases were produced with Claude Opus 4.6 (\texttt{claude-opus-4-6}) using extended thinking (10{,}000-token budget) at temperature 1.0, with the sentence to process passed as the user message and the task specified by the system prompts below. Stage 1 paraphrases the English sentence; Stage 2 translates the English paraphrase into Serbian Cyrillic and Russian Cyrillic. Serbian Latin is not model-generated, but a deterministic 1:1 transliteration of the Serbian Cyrillic. Table~\ref{tab:paraphrase_gen} gives the paraphrase generation pass rate for each of the language-script variants, as well as the mean LaBSE score. 

\paragraph{Stage 1: English Paraphrase Prompt}
\begin{quote}\small
You are a careful paraphrasing assistant. Given an English sentence, generate a paraphrase that:
\begin{enumerate}\itemsep1pt
\item Preserves the exact meaning of the original
\item Uses different vocabulary and sentence structure where natural
\item Sounds like natural, fluent English
\item Maintains roughly similar length (within 30\% of original word count)
\item Keeps proper nouns, numbers, and technical terms exactly as they appear in the original
\item Preserves grammatical gender markers throughout. If the original sentence implies a male or female subject (through pronouns, verb forms, or adjective agreements), the paraphrase must preserve that gender. Do not change ``he'' to ``she'' or vice versa.
\item Preserves the strength and scope of claims. Do not amplify or weaken descriptions:
  \begin{itemize}\itemsep1pt
  \item ``sometimes'' must not become ``often'' or ``always''
  \item ``may'' or ``can'' must not become ``will'' or ``must''
  \item ``outbreak'' must not become ``epidemic''
  \item ``some'' must not become ``most'' or ``all''
  \item Hedged claims must remain hedged
  \item Specific claims must remain specific (do not generalize)
  \end{itemize}
  If the original is uncertain or qualified, the paraphrase must remain qualified to the same degree.
\end{enumerate}
Return ONLY the paraphrased sentence, no preamble, no explanation, no quotes.
\end{quote}

\paragraph{Stage 2a: Serbian Cyrillic Translation}
\begin{quote}\small
You are a careful translator. Translate the given English sentence into natural, fluent Serbian written in Cyrillic script. Preserve the exact meaning. Use natural Serbian phrasing --- do not translate word-for-word from English. Keep proper nouns, numbers, and technical terms in their standard Serbian form (transliterate names if there's a standard Serbian convention, otherwise keep as in English).

For proper nouns, use the standard form of the name in the target language and script. For Serbian Cyrillic, use standard Serbian Cyrillic forms (e.g., Кенеди for Kennedy, Трамп for Trump, Бжежински for Brzezinski). Do not preserve Latin-script spellings within Cyrillic text. However, do not change which person, place, or entity is being referred to: if the original says ``Port Stanley,'' the translation should refer to Port Stanley (in target-language form), not ``Stanley'' or any other place name. Maintain the identity of named entities; localize their surface form to the target script and language conventions.

Preserve grammatical gender markers throughout. If the original sentence implies a male or female subject (through pronouns, verb forms, or adjective agreements), the translation must preserve that gender. In Serbian, maintain consistent gender across all verb forms, pronouns, and adjective endings.

Preserve verb tense and modality exactly:
\begin{itemize}\itemsep1pt
\item Past tense remains past tense
\item Present tense remains present tense
\item Future tense remains future tense
\item Modal verbs (must, may, can, will, should) must map to equivalent strength in Serbian
\item Conditional remains conditional
\item Aspect distinctions (perfective vs imperfective) should match the original's intent
\end{itemize}
Return ONLY the translated sentence, no preamble, no explanation, no quotes.
\end{quote}

\paragraph{Stage 2b: Russian Cyrillic  Translation}
\begin{quote}\small
You are a careful translator. Translate the given English sentence into natural, fluent Russian written in Cyrillic script. Preserve the exact meaning. Use natural Russian phrasing --- do not translate word-for-word from English. Keep proper nouns, numbers, and technical terms in their standard Russian form (transliterate names if there's a standard Russian convention, otherwise keep as in English).

For proper nouns, use the standard form of the name in the target language and script. For Russian, use standard Russian Cyrillic transliterations (e.g., Кеннеди for Kennedy, Трамп for Trump). Do not preserve Latin-script spellings within Cyrillic text. However, do not change which person, place, or entity is being referred to: if the original says ``Port Stanley,'' the translation should refer to Port Stanley (in target-language form), not ``Stanley'' or any other place name. Maintain the identity of named entities; localize their surface form to the target script and language conventions.

Preserve grammatical gender markers throughout. If the original sentence implies a male or female subject (through pronouns, verb forms, or adjective agreements), the translation must preserve that gender. In Russian, maintain consistent gender across all verb forms, pronouns, and adjective endings.

Preserve verb tense and modality exactly:
\begin{itemize}\itemsep1pt
\item Past tense remains past tense
\item Present tense remains present tense
\item Future tense remains future tense
\item Modal verbs (must, may, can, will, should) must map to equivalent strength in Russian
\item Conditional remains conditional
\item Aspect distinctions (perfective vs imperfective) should match the original's intent
\end{itemize}
Return ONLY the translated sentence, no preamble, no explanation, no quotes.
\end{quote}

\begin{table}[h]
\centering
\small
\setlength{\tabcolsep}{4pt}
\begin{tabular}{@{}lrrrr@{}}
\toprule
Stage & $n$ & 1st-try & $\geq$2 att. & LaBSE \\
\midrule
English paraphrase    & 300 & 203 (67.7\%) &  97 (32.3\%) & 0.937 \\
Serbian Cyr.\ trans.  & 300 & 294 (98.0\%) &   6 (2.0\%)  & 0.919 \\
Russian Cyr.\ trans.  & 300 & 296 (98.7\%) &   4 (1.3\%)  & 0.927 \\
Serbian Latin         & 300 & ---          & ---          & 0.919 \\
\bottomrule
\end{tabular}
\caption{Paraphrase generation pass rates on the dataset. ``1st-try'' counts candidates accepted on the first attempt; ``$\geq$2~att.'' counts those that required at least one regeneration to clear the LaBSE $\geq 0.80$ filter. Serbian Latin is produced by deterministic transliteration of Serbian Cyrillic and inherits its acceptance.}
\label{tab:paraphrase_gen}
\end{table}

\subsection{Validation Instructions}
\label{app:validation}

\begin{table*}[t]
\centering
\small
\renewcommand{\arraystretch}{1.3}
\begin{tabularx}{\textwidth}{|>{\raggedright\arraybackslash}X|c|c|>{\raggedright\arraybackslash}X|}
\hline
\textbf{Paraphrase} & \textbf{Meaning} & \textbf{Natural} & \textbf{Why} \\
\hline
On Tuesday, the committee approved the budget by vote. & Yes & Yes & Different wording and structure, same meaning, natural English \\
\hline
The committee voted to reject the budget on Tuesday. & No & Yes & Negation flipped, different meaning \\
\hline
On Tuesday, committee voted approval to budget. & Yes & No & Broken word order, meaning still recoverable \\
\hline
On Tuesday, committee voted to reject approval to budget. & No & No & Meaning reversed (approve to reject) and broken grammar \\
\hline
\end{tabularx}
\caption{Worked examples shown to validators, illustrating the two-axis judgment. Original sentence: \emph{``The committee voted to approve the budget on Tuesday.''} Examples were presented in each validator's target language; English is shown here for readability.}
\label{tab:validation-examples}
\end{table*}

Native-speaker validators received identical instructions for each language, with the target language substituted throughout. Each validator saw sentence pairs, an original sentence and its candidate paraphrase in the same language and script, and rated each pair on two binary axes. Table~\ref{tab:validation-examples} shows worked examples of the four Meaning and Naturalness combinations.

\paragraph{Meaning.} \emph{Yes} if both sentences make the same factual claim, such that a reader of either would understand the same thing; \emph{No} if the paraphrase adds, removes, or changes the claim. Validators were told not to mark \emph{No} merely for differences in wording, word order, or style, since paraphrases are expected to differ on the surface. They were instructed to mark Meaning \emph{No} whenever any of the following occurred: a flipped negation; a meaning-relevant change of tense or aspect; a changed number; a changed or omitted named entity (person, place, organization, or date); a changed modality; a reversed comparison; a swapped subject and object; or any other omission, weakening, or addition that alters the claim. 

\paragraph{Naturalness.} \emph{Yes} if the paraphrase reads as fluent text a native speaker could plausibly have written; \emph{No} if a native reader would notice something off without being told the text was machine-generated. Validators were told they did not need to name a grammatical rule; a native-speaker judgment that the text sounded wrong was sufficient. Mild shifts in formality were not grounds for \emph{No} unless the register itself sounded unnatural.

\paragraph{Corrections.} A pair rated \emph{Yes} on both axes was accepted as is. Otherwise the validator supplied a corrected paraphrase, written in the target script, that preserved the original meaning and read naturally, along with a short note describing the change. If a paraphrase was too broken to judge confidently, validators marked both axes \emph{No} and supplied a correction. Corrected paraphrases enter the dataset flagged separately from accepted model paraphrases, so the two can always be told apart, and no items are discarded.

\paragraph{Compensation.} Each validator was paid \$125--\$200 per batch of
200 sentence pairs, recruited through personal networks as native speakers of
the target language. Payment was calibrated to exceed standard crowdsourcing
rates for comparable linguistic annotation tasks. Before starting, validators were informed that their judgments and corrections
would be released as part of a public research dataset, and agreed to
participate on that basis.

\subsection{Disagreement Resolution}
\label{app:resolution}

For the 100 sentences in each language judged by both validators, a single decision per item populates the dataset, resolved conservatively. Table~\ref{tab:disagreement} gives the per-language breakdown. When neither validator flagged the pair, the model paraphrase was kept. When exactly one validator flagged it, we adopted that validator's correction. When both flagged the same item with differing corrections, we kept the correction from one designated validator when the sentence's FLORES+ index was even and from the other when it was odd. Because the index is assigned by the benchmark before validation, this rule was fixed in advance and cannot depend on the content or favor either validator. One-sided disagreements (0, 11, and 23 items for English, Russian, and Serbian) were therefore resolved deterministically, and the parity tiebreak applied only to items both validators flagged with differing corrections, at most 1, 2, and 14 items respectively, each a choice between two validator-approved paraphrases. 

\begin{table}[t]
\centering
\small
\renewcommand{\arraystretch}{1.3}
\begin{tabular}{|l|c|c|c|}
\hline
\textbf{Overlap outcome} & \textbf{English} & \textbf{Russian} & \textbf{Serbian} \\
\hline
Neither flagged & 99 & 87 & 63 \\
\hline
Exactly one flagged & 0 & 11 & 23 \\
\hline
Both flagged & 1 & 2 & 14 \\
\hline
\end{tabular}
\caption{Validator decisions on the 100-sentence overlap subset, by accept-or-flag outcome (rows sum to 100). Items neither validator flagged kept the model paraphrase; when only one validator flagged a pair, we adopted that validator's correction; items both validators flagged used the parity tiebreak only when the two corrections differed.}
\label{tab:disagreement}
\end{table}



\subsection{Detailed Validation Results}
\label{app:valdetail}

This section reports the full per-language validation statistics summarized in the main text. Table~\ref{tab:val-outcomes} gives clean-accept and correction rates; the lower Serbian rate reflects rougher generated Serbian, which validators caught and corrected rather than discarded. Table~\ref{tab:val-failtype} breaks those corrections down by what failed: English corrections were almost all meaning fixes, every Russian correction involved naturalness, and Serbian spanned both. Table~\ref{tab:val-edit} reports how large those corrections were, as the word-level edit distance between each model paraphrase and the validator's replacement; the median touches one to three words in every language, and only Serbian shows a tail of larger rewrites, visible as a mean fraction well above its median. Table~\ref{tab:val-irr} reports inter-rater agreement on the 100-sentence overlap subset, measuring how often the two validators for a language reached the same accept-or-correct decision on the items they both judged.  We rely on raw agreement and Gwet's AC1 rather than Cohen's $\kappa$, since the high accept rates produce skewed marginals under which $\kappa$ is deflated even when raw agreement is high. Agreement is perfect for English, high for Russian, and moderate for Serbian. For English the two validators made the same accept-or-correct decision on all 100 overlap items: the only item that drew a correction was flagged by both, and both supplied the same fix, so raw agreement and AC1 are both 1.0. For Serbian, validators differed mainly on whether a borderline paraphrase warranted a minor fix, not on whether meaning was preserved.

\begin{table}[H]
\centering
\small
\begin{tabular}{@{}lccc@{}}
\toprule
\textbf{Metric} & \textbf{English} & \textbf{Serbian} & \textbf{Russian} \\
\midrule
Clean-accept & 288 (96.0\%) & 212 (70.7\%) & 272 (90.7\%) \\
Correction   & 12 (4.0\%)   & 88 (29.3\%)  & 28 (9.3\%) \\
\bottomrule
\end{tabular}
\caption{Per-language outcomes over 300 sentences (counts out of 300). Serbian Cyrillic and Latin share these rates,
since Latin inherits Cyrillic corrections via transliteration. The
correction counts (12, 88, 28) match the totals in
Table~\ref{tab:val-failtype}. }
\label{tab:val-outcomes}
\end{table}

\begin{table}[h]
\centering
\small
\setlength{\tabcolsep}{4pt}
\begin{tabular}{@{}lccc@{}}
\toprule
\textbf{Failure Type} & \textbf{English} & \textbf{Serbian} & \textbf{Russian} \\
\midrule
Meaning Incorrect     & 11 (91.7\%) & 57 (64.8\%) & 0 (0\%) \\
Naturalness Incorrect & 1 (8.3\%)   & 14 (15.9\%) & 18 (64.3\%) \\
Both             & 0 (0\%)     & 17 (19.3\%) & 10 (35.7\%) \\
\midrule
Total  & 12 & 88 & 28 \\
\bottomrule
\end{tabular}
\caption{Reason for correction, by language, over corrected paraphrases only.
The three categories are mutually exclusive: each corrected item is counted
once, by whether it failed on meaning, on naturalness, or on both.}
\label{tab:val-failtype}
\end{table}

\begin{table}[t]
\centering
\small
\renewcommand{\arraystretch}{1.3}
\begin{tabular}{|l|c|c|c|c|}
\hline
\textbf{Metric} & \textbf{Eng} & \textbf{Srp-Cyrl} & \textbf{Srp-Latn} & \textbf{Rus} \\
\hline
Median edit & 1.0 & 2.5 & 2.5 & 2.0 \\
\hline
Mean edit & 2.17 & 3.25 & 3.25 & 3.79 \\
\hline
Median frac. & 0.077 & 0.122 & 0.122 & 0.131 \\
\hline
Mean frac. & 0.105 & 0.433 & 0.433 & 0.213 \\
\hline
\end{tabular}
\caption{Edit magnitude over corrected items only, Edit distances are word-level; ``frac.'' is that distance as a fraction of sentence length.  The Serbian mean fraction is inflated by a small tail of larger rewrites, while the median stays low.}
\label{tab:val-edit}
\end{table}

\begin{table}[t]
\centering
\small
\renewcommand{\arraystretch}{1.3}
\begin{tabular}{|l|c|c|c|}
\hline
\textbf{Metric} & \textbf{English} & \textbf{Russian} & \textbf{Serbian} \\
\hline
Raw agreement & 100\% & 89\% & 77\% \\
\hline
Cohen's $\kappa$ & n/a & 0.21 & 0.40 \\
\hline
Gwet's AC1 & 1.0 & 0.87 & 0.63 \\
\hline
\end{tabular}
\caption{Inter-rater agreement on the 100-sentence overlap subset. }
\label{tab:val-irr}
\end{table}

\subsection{Length and Tokenization}
\label{app:lentok}

We report sentence length in words and in token count, both to
characterize the conditions and to address the concern that cross-script
effects could reflect tokenization rather than representation.
Table~\ref{tab:lentok-counts} gives both counts by condition and variant.
Word counts are matched across the three conditions: originals,
paraphrases, and random partners average $20.2$, $20.9$, and $20.2$ words
respectively, with paraphrases running under one word longer than their
originals, well inside the 30\% length tolerance. Variants differ
modestly in word count (English averages about one word more than Serbian,
Russian about one word less), reflecting language-level differences in how
the same content is expressed rather than any condition effect.

\begin{table*}[t]
\centering
\small
\setlength{\tabcolsep}{5pt}
\begin{tabular}{lccccc}
\toprule
\textbf{Condition} & \textbf{En-Latin} & \textbf{Sr-Cyrillic} & \textbf{Sr-Latin} & \textbf{Ru-Cyrillic} & \textbf{Marginal} \\
\midrule
\multicolumn{6}{l}{\emph{Word Count}} \\
Original   & $21.4 \pm 6.7$ (20) & $20.1 \pm 6.6$ (19) & $20.1 \pm 6.6$ (19) & $19.2 \pm 6.5$ (18) & $20.2 \pm 6.6$ (19) \\
Paraphrase & $22.3 \pm 6.9$ (22) & $20.8 \pm 6.8$ (20) & $20.8 \pm 6.8$ (20) & $19.8 \pm 6.6$ (19) & $20.9 \pm 6.8$ (20) \\
Random     & $21.3 \pm 6.4$ (21) & $20.1 \pm 6.3$ (20) & $20.1 \pm 6.3$ (20) & $19.3 \pm 6.2$ (19) & $20.2 \pm 6.3$ (20) \\
\midrule
\multicolumn{6}{l}{\emph{Token Count}} \\
Original   & $26.3 \pm 8.6$ (25) & $45.0 \pm 14.8$ (42) & $39.5 \pm 13.2$ (37) & $35.8 \pm 12.4$ (34) & $36.7 \pm 14.2$ (35) \\
Paraphrase & $27.4 \pm 8.9$ (26) & $47.9 \pm 15.3$ (46) & $41.8 \pm 13.5$ (39) & $37.8 \pm 12.7$ (36) & $38.7 \pm 14.8$ (36) \\
Random     & $26.6 \pm 8.3$ (26) & $46.1 \pm 14.2$ (45) & $40.1 \pm 12.4$ (39) & $36.2 \pm 12.0$ (36) & $37.3 \pm 13.9$ (36) \\
\bottomrule
\end{tabular}
\caption{Word and subword token counts by condition and language-script
variant, as mean $\pm$ standard deviation with median in parentheses. Word
counts are matched across conditions; subword counts differ sharply across
scripts. Token counts use the Gemma-3-27B tokenizer.}
\label{tab:lentok-counts}
\end{table*}

\begin{table}[t]
\centering
\small
\begin{tabular}{lcccc}
\toprule
\textbf{Condition} & \textbf{En-Lat} & \textbf{Sr-Cyr} & \textbf{Sr-Lat} & \textbf{Ru-Cyr} \\
\midrule
Original   & 1.24 & 2.26 & 1.98 & 1.88 \\
Paraphrase & 1.23 & 2.33 & 2.03 & 1.93 \\
Random     & 1.25 & 2.31 & 2.01 & 1.90 \\
\bottomrule
\end{tabular}
\caption{Tokens per word (Gemma-3-27B). The same content costs roughly twice
as many tokens in Serbian Cyrillic as in English, and about $15\%$ more in
Serbian Cyrillic than in the deterministically transliterated Serbian Latin.}
\label{tab:tokratio}
\end{table}

Subword tokenization, by contrast, differs sharply across scripts.
Table~\ref{tab:tokratio} reports tokens per word under the Gemma-3-27B
tokenizer. English costs about $1.2$ tokens per word, whereas Serbian
Cyrillic costs about $2.3$, so the same roughly $20$-word content expands
from about $26$ tokens in English to $45$--$48$ in Serbian Cyrillic. The
effect is partly script and partly language: Serbian Latin, a deterministic
transliteration of the identical Serbian Cyrillic content, costs about $2.0$
tokens per word, so the Cyrillic script alone adds roughly $15\%$ more tokens
for word-for-word identical text. Russian Cyrillic is more efficient than
Serbian Cyrillic (about $1.9$ versus $2.3$).

These large, script-driven token-count differences are intrinsic to the
comparison and cannot be removed without distorting content: equalizing
token counts across scripts would require unmatching the content itself. Our
design therefore controls content length in words and treats subword
tokenization as a property of each variant. This raises the question of
whether these token-count differences themselves drive the feature overlap
patterns we report.

\begin{figure}[h]
\centering
\includegraphics[width=\columnwidth]{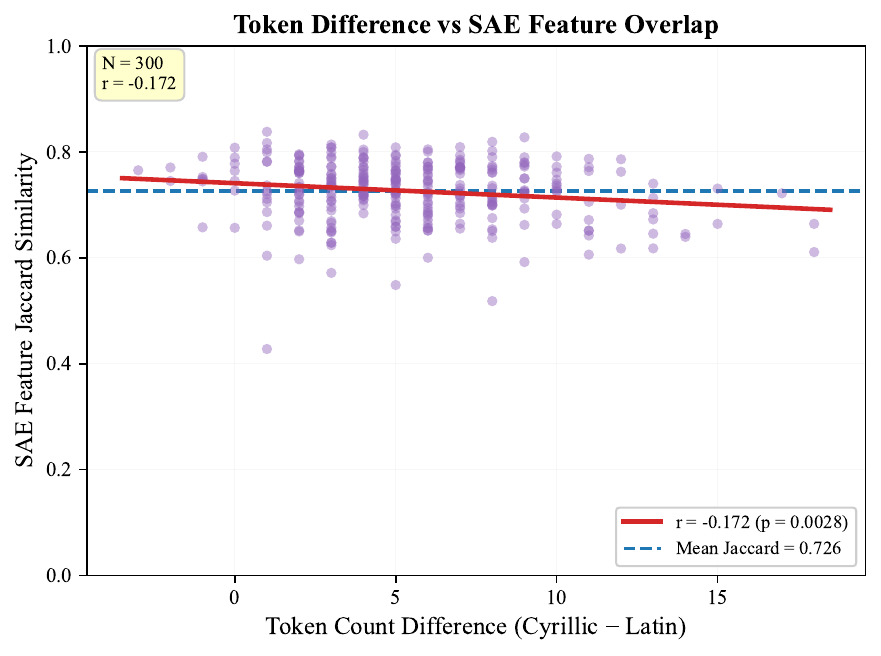}
\caption{Cross-script Jaccard similarity (averaged across layers) vs.\
token count difference (Cyrillic minus Latin) for 300 Serbian sentence
pairs. The regression line is nearly flat ($r = -0.17$, $r^2 = 0.03$).}
\label{fig:token_confound}
\end{figure}

\paragraph{Tokenization and feature overlap.}
We test this directly. 
For each of the 300 Serbian sentence pairs (same sentence in Latin and
Cyrillic), we compute the token count difference (Cyrillic minus Latin)
and correlate it with the cross-script Jaccard similarity averaged
across all layers. Token deltas ranged from $-3$ to $19$ (mean $5.4$,
sd $3.6$, IQR $[3, 8]$), confirming that Cyrillic consistently
tokenizes into more subwords. The correlation with feature overlap is
statistically significant but negligible ($r = -0.172$, $p = 0.003$,
$r^2 = 0.03$): tokenization mismatch accounts for roughly 3\% of the
variance in Jaccard similarity. Active feature set sizes show no
relationship with token count differences at all ($r = 0.027$,
$p = 0.639$), ruling out a mechanism in which longer tokenizations
produce systematically different feature counts. These results
indicate that tokenization asymmetry is not a meaningful driver of the
cross-script patterns we report. Figure~\ref{fig:token_confound} shows
the relationship visually.

\subsection{Active-Feature Activation Magnitudes}
\label{app:magnitude}

The active set $F_\ell(s)$ is defined by nonzero post-JumpReLU activation. Because JumpReLU emits zero below a learned per-feature threshold $\theta$ and a value of at least $\theta$ above it, ``nonzero'' is not a near-zero cutoff but a substantial gate. We verify that active features clear this gate by a comfortable margin rather than skimming it. Table~\ref{tab:magnitude} reports activation statistics at layer~31 on Gemma-3-27B (illustrative; layer~31 lies in the mid-network region where the meaning signal is high).

Active features fire well above their threshold: the median active feature activates at $1.67\times$ its gate $\theta$, and only $11.6\%$ of active features sit within $10\%$ of $\theta$. The binary $>0$ criterion therefore selects genuinely engaged features rather than threshold noise, supporting its use as the basis for our feature-set analyses.

\begin{table}[t]
\centering
\small
\begin{tabular}{@{}lr@{}}
\toprule
Quantity & Value \\
\midrule
Active (feature, text) pairs            & 80{,}578 \\
Active features per text (mean)         & 22.4 \\
Activation magnitude (median)           & 518.4 \\
Activation magnitude (mean)             & 714.2 \\
Activation magnitude (p5 / p95)         & 319.4 / 1{,}831.9 \\
Threshold $\theta$ (min / median / max) & 201.2 / 339.2 / 5{,}197.6 \\
Median margin over $\theta$             & $1.67\times$ \\
Active features within $10\%$ of $\theta$ & $11.6\%$ \\
\bottomrule
\end{tabular}
\caption{Active-feature activation magnitudes at layer~31 (Gemma-3-27B). Active features fire well above the JumpReLU threshold, confirming the $>0$ active-set criterion captures genuine engagement.}
\label{tab:magnitude}
\end{table}

\subsection{Dataset Composition by Source and Topic}
\label{app:domain}
Source-domain composition is given in Table~\ref{tab:domain}. Originals and
paraphrases preserve the FLORES+ stratification exactly; random partners are
sampled within language-script and length-matched. Table~\ref{tab:topic-breakdown} gives the finer-grained topic
distribution; the dataset spans a broad range of subject matter, with no single
topic dominating.
\begin{table}[t]
\centering
\small
\begin{tabular}{lccc}
\toprule
\textbf{Domain} & \textbf{Orig.} & \textbf{Para.} & \textbf{Rand.} \\
\midrule
Wikinews   & 100 & 100 & 105 \\
Wikibooks  & 101 & 101 & 102 \\
Wikivoyage &  99 &  99 &  93 \\
\bottomrule
\end{tabular}
\caption{Source-domain counts over the $300$ anchors. Originals and
paraphrases preserve the FLORES+ stratification exactly; random partners are
sampled within language-script and length-matched, shifting domain
composition slightly.}
\label{tab:domain}
\end{table}

\begin{table}[t]
\centering
\footnotesize
\renewcommand{\arraystretch}{1.0}
\setlength{\tabcolsep}{4pt}
\begin{tabular}{lrrr}
\toprule
\textbf{Topic} & \textbf{Count} & \textbf{\%} & \textbf{Primary Domain} \\
\midrule
Travel \& Tourism & 91 & 30.3 & Wikivoyage \\
Science \& Technology & 79 & 26.3 & Wikibooks \\
History \& Geography & 50 & 16.7 & Wikibooks \\
Politics \& Law & 33 & 11.0 & Wikinews \\
Culture \& Entertainment & 27 & 9.0 & Wikinews \\
Sports & 22 & 7.3 & Wikinews \\
Education \& Social Science & 19 & 6.3 & Wikibooks \\
Disasters \& Accidents & 11 & 3.7 & Wikinews \\
Other & 1 & 0.3 & Wikinews \\
\bottomrule
\end{tabular}
\caption{Topic distribution of the 300 original sentences, grouped into broad categories from FLORES+ metadata. Percentages exceed 100\% because some sentences carry multiple topic tags (e.g., ``disease, research, canada'').}
\label{tab:topic-breakdown}
\end{table}

\clearpage
\section{Ablation Testing}
\label{app:results}

\subsection{SAE Sparsity Level}
\label{app:sparsity}

\begin{figure*}[t]
\centering
\includegraphics[width=\textwidth]{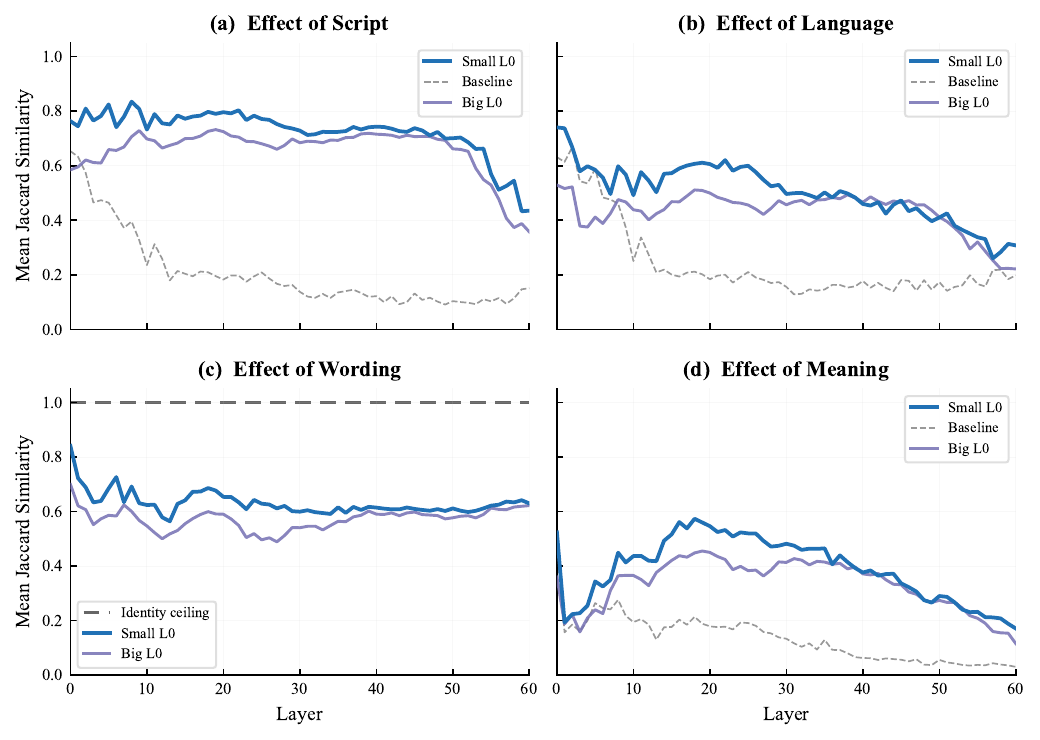}
\caption{Leg~1 decomposition comparing small-$L_0$ (blue) and large-$L_0$
(purple) Gemma Scope~2 SAEs at 16K width on Gemma-3-27B. Dashed grey lines
show baselines. The factorial ordering and depth structure are preserved.}
\label{fig:sparsity}
\end{figure*}

We compare the small-$L_0$ SAEs used in the main text with the large-$L_0$
variant from Gemma Scope~2, both at 16K width
(Figure~\ref{fig:sparsity}). Higher $L_0$ means more features fire per
input, producing larger and noisier active sets.  The decomposition hierarchy and layer-wise trajectories are fully preserved, confirming that the results are not sensitive to sparsity level.

\subsection{SAE Dictionary Width}
\label{app:width}

\begin{figure*}[t]
\centering
\includegraphics[width=\textwidth]{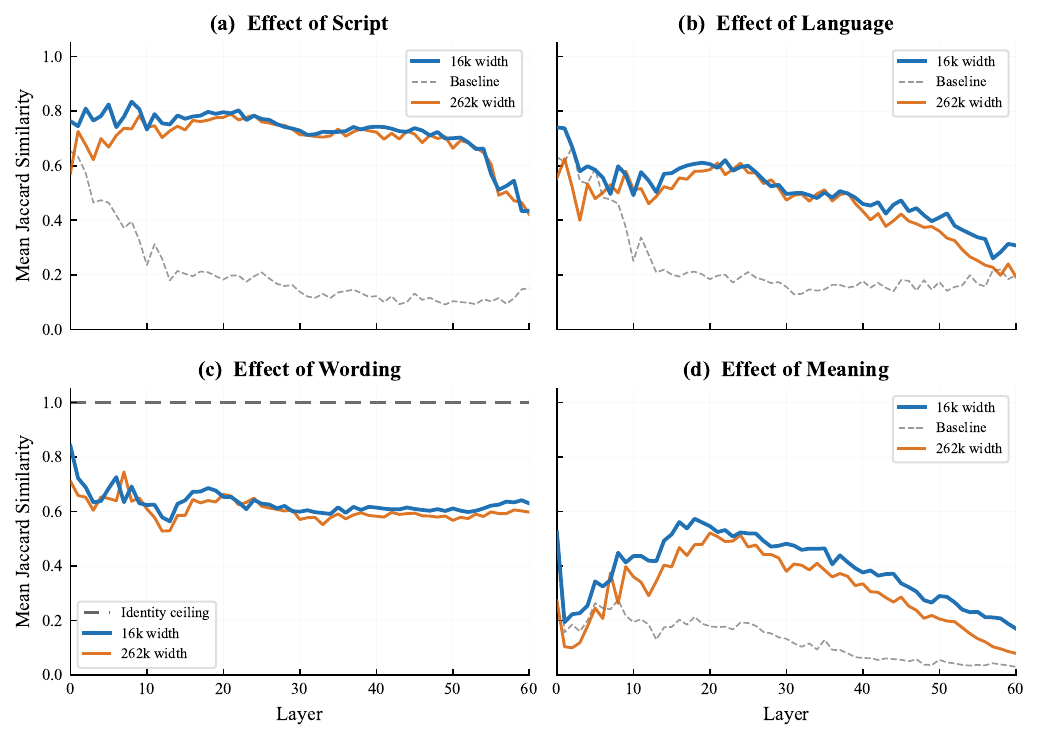}
\caption{Leg~1 decomposition comparing 16K-width (blue) and 262K-width
(orange) Gemma Scope~2 SAEs on Gemma-3-27B. Dashed grey lines show baselines.
The layer-wise shape and panel ordering are preserved; the uniform downward
shift reflects sparser feature activation in the wider dictionary.}
\label{fig:width}
\end{figure*}

We replicate the Leg~1 decomposition on Gemma-3-27B using 262K-width Gemma
Scope~2 SAEs, a $16\times$ increase over the primary 16K configuration
(Figure~\ref{fig:width}). The layer-wise trajectories and the
ordering across panels are nearly identical to the 16K results, confirming
that the factorial decomposition is not an artifact of dictionary size.

\subsection{Pooling Strategy}
\label{app:pooling}

\begin{figure*}[t]
\centering
\includegraphics[width=\textwidth]{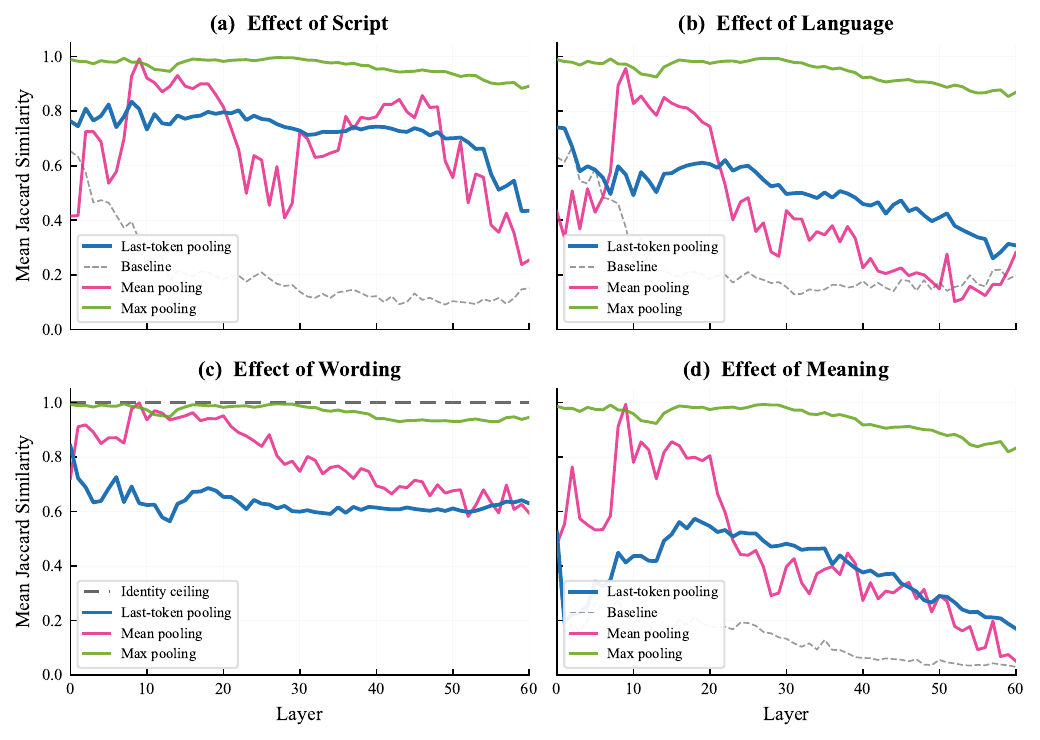}
\caption{Pooling strategy comparison on Gemma-3-27B. Dashed grey lines show
the corresponding baselines. }
\label{fig:pooling}
\end{figure*}

We compare last-token, mean, and max pooling on Gemma-3-27B
(Figure~\ref{fig:pooling}). The factorial ordering (script $>$ wording $>$ language
$>$ meaning) holds under both last-token and mean pooling; only the effect
sizes differ. Among last-token, mean, and max pooling, only last-token cleanly separates the conditions from their baselines; max saturates near 1.0 and mean is erratic across layers, so we adopt last-token.

\subsection{Model Scale}
\label{app:scale}

\begin{figure*}[t]
\centering
\includegraphics[width=\textwidth]{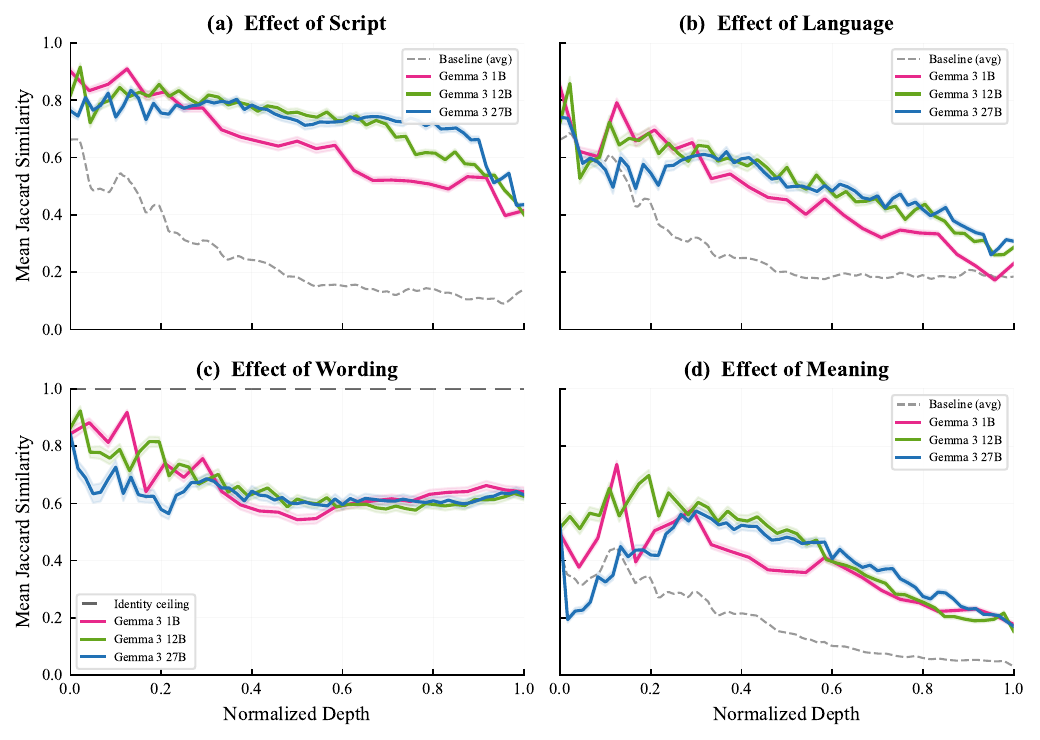}
\caption{Leg~1 decomposition across Gemma-3 model sizes (1B, 12B, 27B),
plotted on normalized depth (300 sentences, 95\% bootstrap CIs). Solid lines
show the main comparison for each panel; thin grey dashed lines show the
corresponding random-partner baseline averaged across all three model sizes
(per-model baselines are nearly identical). Panel~(c) retains the identity
ceiling at 1.0. The qualitative structure holds at all three scales: script
overlap remains highest, meaning stays well above its baseline, and wording is
roughly flat. }
\label{fig:scale}
\end{figure*}

To verify that our findings are not specific to Gemma-3-27B, we replicate the
full Leg~1 factorial decomposition on Gemma-3-1B (25 layers) and Gemma-3-12B
(47 layers) using the same dataset, SAE configuration (16k width, low L0), and
pipeline. Figure~\ref{fig:scale} overlays the four panels across all three
model sizes on a normalized depth axis.

The core structure holds across scale: in all three models, script overlap
exceeds language overlap, meaning rises well above its random baseline, and
wording stays roughly flat. The depth trajectories
are noisier at 1B, where the model has fewer layers to differentiate, but the
qualitative ordering of the four panels is preserved. We conclude that the
factorial decomposition reported in the main text is robust to model scale
within the Gemma family.

\subsection{Cross-Architecture: Llama-3.1-8B}
\label{app:llama}

To verify that our findings are not specific to the Gemma architecture or
Gemma Scope 2 SAEs, we replicate the Leg~1 decomposition on
Llama-3.1-8B-Base using Llama Scope SAEs \citep{he2024llamascope}: 32K-width
(8$\times$ expansion) post-MLP residual stream SAEs at all 32 layers. These
SAEs are trained with TopK activation ($k{=}50$) and decoder-norm-weighted
selection, then converted to JumpReLU with per-layer thresholds for
independent feature activation at inference; we define the active set as
features with nonzero post-JumpReLU activation. This setup differs from our
primary analysis in model family, SAE training regime, dictionary width
(32K vs.\ 16K), and tokenizer.

\begin{figure*}[t]
\centering
\includegraphics[width=\textwidth]{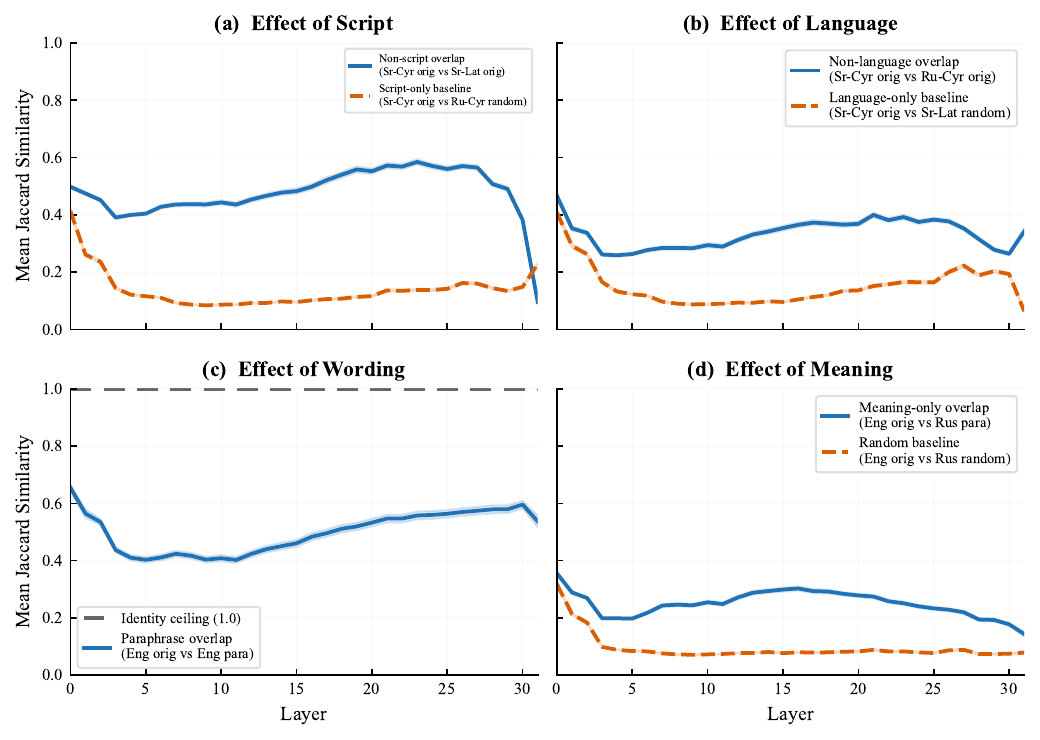}
\caption{Leg~1 decomposition for Llama-3.1-8B-Base with Llama Scope SAEs
(32K width, 32 layers, 300 sentences). Solid lines show the main comparison;
dashed lines show random-partner baselines. Despite a different model family,
SAE training regime (TopK-trained, JumpReLU-converted), and dictionary width
(32K vs.\ 16K), the qualitative structure matches Gemma-3-27B
(Figure~\ref{fig:decomposition}): script overlap highest, meaning above
baseline peaking mid-network, wording roughly flat.}
\label{fig:llama}
\end{figure*}

The qualitative structure replicates (Figure~\ref{fig:llama}). Script overlap
is highest and dips in late layers, meaning peaks in the middle of the network
and stays well above its random baseline, and wording remains in the
$0.4$--$0.6$ range throughout. Absolute Jaccard values are lower across the
board (e.g.\ script mean $0.48$ vs.\ $0.73$ for Gemma-3-27B), but the
ordering across panels and the depth trajectories are preserved. The
factorial decomposition is therefore not an artifact of the Gemma architecture
or Gemma Scope 2 SAEs.

\clearpage
{\raggedbottom
\section{Robustness-Check Classifier}
\label{app:classifier}
This appendix details the secondary content classifier and its per-layer discard rates.

\paragraph{Classifier Configs.}
Every retrieved auto-interp label is passed through a secondary classifier built on \texttt{claude-sonnet-4-6} (\texttt{max\_tokens}=2000). Each feature is assigned to exactly one of five categories: \textit{content-claim}, \textit{surface-claim}, \textit{language-claim}, \textit{mixed}, or \textit{other}; any response outside this set is coerced to \textit{other}. Only \textit{content-claim} features propagate into the downstream falsification, miss-rate, asymmetry, and typology analyses. The exact prompt is shown below.

\paragraph{Classifier Prompt.}\mbox{}\\[2pt]
{\footnotesize\ttfamily\setlength{\parindent}{0pt}%
Classify each SAE feature auto-interp label\\
into exactly one category:\\[3pt]
content-claim: semantic content, topic,\\
\hspace*{1em}or meaning\\
surface-claim: orthography, script,\\
\hspace*{1em}formatting, or specific tokens\\
language-claim: a particular language\\
\hspace*{1em}or family\\
mixed: content + surface/language\\
other: doesn't fit the above\\[3pt]
Labels (feature\_id: label): \{label\_block\}\\
Respond with ONLY one line per label:\\
\hspace*{1em}feature\_id: category.\par}

\paragraph{Discard Rates.}
Tables~\ref{tab:discard_sonnet_27b} and~\ref{tab:discard_gemini_27b} report the per-layer outcome of the content-claim gate on Gemma~3 27B. Sonnet discards 1{,}744 of 5{,}238 labels (33.3\%); Gemini discards 1{,}401 of 5{,}211 (26.9\%). Both attribute the bulk of the drop to \textit{surface-claim} labels (Tables~\ref{tab:discard_sonnet_breakdown_27b} and~\ref{tab:discard_gemini_breakdown_27b}). The comparison in Table~\ref{tab:discard_sonnet_vs_gemini_27b} shows Sonnet is stricter at every layer, with the gap widening sharply at layer~53 (45.2\% vs.\ 29.7\%). Despite these differing discard rates, both labelers converge on the same Cyrillic/Latin asymmetry at every layer (Table~\ref{tab:27b_ratio_featureCI}), confirming the asymmetry is a property of the features rather than of either classifier's strictness.

\begin{table}[H]
\centering
\small
\setlength{\tabcolsep}{4pt}
\begin{tabular}{@{}lrrrr@{}}
\toprule
Layer & Labels & Content & Disc. & \% Disc. \\
\midrule
16    &   852 &   596 & 256 & 30.0 \\
31    & 1{,}213 &   896 & 317 & 26.1 \\
40    & 1{,}447 & 1{,}056 & 391 & 27.0 \\
53    & 1{,}726 &   946 & 780 & 45.2 \\
\midrule
\textbf{Total} & \textbf{5{,}238} & \textbf{3{,}494} & \textbf{1{,}744} & \textbf{33.3} \\
\bottomrule
\end{tabular}
\caption{Sonnet classifier discards on Gemma~3 27B (leg~1) across the four Neuronpedia SAE layers. ``Disc.'' counts any label not tagged \textit{content-claim}.}
\label{tab:discard_sonnet_27b}
\end{table}

\begin{table}[H]
\centering
\small
\setlength{\tabcolsep}{4pt}
\begin{tabular}{@{}lr@{}}
\toprule
Category & Count \\
\midrule
surface-claim         & 1{,}413 \\
mixed                 &    172 \\
language-claim        &     79 \\
other                 &     78 \\
unknown (parse fail)  &      2 \\
\midrule
\textbf{Total non-content} & \textbf{1{,}744} \\
\bottomrule
\end{tabular}
\caption{Composition of the 1{,}744 Sonnet-classifier discards on Gemma~3 27B, summed across all four layers.}
\label{tab:discard_sonnet_breakdown_27b}
\end{table}

\begin{table}[H]
\centering
\small
\setlength{\tabcolsep}{4pt}
\begin{tabular}{@{}lrrrr@{}}
\toprule
Layer & Labels & Content & Disc. & \% Disc. \\
\midrule
16    &   840 &   592 & 248 & 29.5 \\
31    & 1{,}219 &   883 & 336 & 27.6 \\
40    & 1{,}436 & 1{,}128 & 308 & 21.4 \\
53    & 1{,}716 & 1{,}207 & 509 & 29.7 \\
\midrule
\textbf{Total} & \textbf{5{,}211} & \textbf{3{,}810} & \textbf{1{,}401} & \textbf{26.9} \\
\bottomrule
\end{tabular}
\caption{Gemini classifier discards on Gemma~3 27B (leg~1) across the four Neuronpedia SAE layers, using the same content-claim gate.}
\label{tab:discard_gemini_27b}
\end{table}

\begin{table}[H]
\centering
\small
\setlength{\tabcolsep}{4pt}
\begin{tabular}{@{}lr@{}}
\toprule
Category & Count \\
\midrule
surface-claim         & 968 \\
mixed                 & 218 \\
other                 & 154 \\
language-claim        &  53 \\
unknown (parse fail)  &   8 \\
\midrule
\textbf{Total non-content} & \textbf{1{,}401} \\
\bottomrule
\end{tabular}
\caption{Composition of the 1{,}401 Gemini-classifier discards on Gemma~3 27B, summed across all four layers.}
\label{tab:discard_gemini_breakdown_27b}
\end{table}

\begin{table}[H]
\centering
\small
\setlength{\tabcolsep}{4pt}
\begin{tabular}{@{}lcc@{}}
\toprule
Layer & Sonnet \% & Gemini \% \\
\midrule
16    & 30.0 & 29.5 \\
31    & 26.1 & 27.6 \\
40    & 27.0 & 21.4 \\
53    & 45.2 & 29.7 \\
\midrule
\textbf{Overall} & \textbf{33.3} & \textbf{26.9} \\
\bottomrule
\end{tabular}
\caption{Cross-classifier comparison of discard rates on Gemma~3 27B. Sonnet is the stricter judge at every layer, with the gap widening sharply at layer~53.}
\label{tab:discard_sonnet_vs_gemini_27b}
\end{table}
}

\clearpage
\section{Detailed Miss-Rate Breakdowns}
\label{app:detailedbreakdown}

Tables~\ref{tab:appendix-sonnet-maxvar} and~\ref{tab:appendix-gemini-maxvar}
report the full miss-rate breakdown for the cross-lingual content pool under Sonnet
and Gemini labels, respectively. Each table includes within-language
paraphrase miss rates (Eng-orig, Rus-para), cross-language Serbian miss rates
for both originals and paraphrases, and random baselines confirming
content-selectivity (77--96\% miss on controlled-unrelated sentences).
Table~\ref{tab:labeler-comparison-maxvar} compares the two labelers side by
side: despite classifying different subsets as content-claim, both converge on
the same Cyrillic/Latin asymmetry ratio at every layer (e.g., 1.35$\times$
vs.\ 1.36$\times$ at layer 53). Table~\ref{tab:vignettes-appendix} provides
20 additional feature vignettes beyond the three in the main text, spanning
all four SAE layers and both failure directions, with full English sentences
and dual-labeler agreement. Tables~\ref{tab:per-topic-asymmetry} and~\ref{tab:per-topic-asymmetry-multi} further show that the Cyrillic disadvantage holds within every FLORES+ topic bucket under both primary- and multi-tag assignment. Finally, to confirm the content classifier itself does not drive the asymmetry, Table~\ref{tab:27b_ratio_featureCI} reports miss-rate ratios computed on the \emph{unfiltered} pool, with no content gate applied. The content filter exists to restrict the analysis to features making genuine content claims, excluding surface- and language-claims (e.g., 'Cyrillic characters', 'Russian morphology') that would differ across scripts by construction and inflate the asymmetry independently of any content-generalization effect. That the asymmetry nonetheless holds in the unfiltered pool, with ratios overlapping the content-filtered version at every layer (excluding 1.0 from layer 31 onward), shows the filter sharpens the interpretation of the result rather than producing the cross-script gap.

\subsection{Reproducibility}
Auto-interp labels were generated by Claude Sonnet 4.5  and by Gemini 2.5 Flash Lite (gemini-2.5-flash-lite, Neuronpedia's default labeler). Neuronpedia itself exposes no API-level version, snapshot, or per-explanation timestamp, and served labels are mutable. All Neuronpedia GETs were issued between 2026-05-24 and 2026-05-25 (UTC-local), during which the deployed Neuronpedia codebase was release v1.0.755. To enable byte-exact replication independent of future label regenerations or model deprecations, we release the cached Neuronpedia and Sonnet JSON bundles alongside our code; users should verify against these snapshots rather than live Neuronpedia.

\begin{table*}[t]
\centering
\small
\setlength{\tabcolsep}{3.5pt}
\begin{tabular}{@{}lcccccccc@{}}
\toprule
 & \multicolumn{2}{c}{Within-language (\%)} & \multicolumn{2}{c}{Serbian orig (\%)} & \multicolumn{2}{c}{Serbian para (\%)} & \multicolumn{2}{c}{Cyr/Lat ratio} \\
\cmidrule(lr){2-3} \cmidrule(lr){4-5} \cmidrule(lr){6-7} \cmidrule(lr){8-9}
Layer & Eng-para & Rus-orig & Sr-Latin & Sr-Cyrillic & Sr-Latin & Sr-Cyrillic & Orig & Para \\
\midrule
16 & 14.8 [12.5, 18.4] & 16.6 [12.5, 22.3] & 20.8 [15.9, 27.4] & 21.3 [15.8, 28.8] & 18.2 [13.9, 24.3] & 19.1 [13.8, 26.4] & 1.02$\times$ & 1.05$\times$ \\
31 &  9.2 [7.0, 11.6]  & 13.7 [11.4, 16.6] & 20.4 [18.1, 23.4] & 22.6 [20.3, 25.1] & 18.1 [15.9, 20.8] & 21.8 [18.9, 24.7] & 1.11$\times$ & 1.21$\times$ \\
40 &  9.1 [7.7, 10.5]  & 15.6 [14.0, 17.3] & 22.2 [19.9, 24.6] & 25.9 [23.4, 28.6] & 20.7 [18.7, 22.9] & 25.2 [22.6, 27.9] & 1.17$\times$ & 1.22$\times$ \\
53 &  7.3 [5.8, 9.0]   & 15.4 [13.4, 17.5] & 25.6 [22.1, 29.6] & 34.6 [30.7, 38.6] & 24.8 [20.6, 29.2] & 33.6 [29.4, 38.1] & 1.35$\times$ & 1.36$\times$ \\
\midrule
 & \multicolumn{4}{c}{Random baselines (\%)} \\
\cmidrule(lr){2-5}
Layer & Eng & Rus & Sr-Latin & Sr-Cyrillic \\
\cmidrule(lr){2-5}
16 & 78.3 [67.7, 92.2] & 76.8 [65.9, 91.9] & 77.4 [65.6, 92.9] & 77.6 [66.1, 92.7] \\
31 & 81.3 [66.1, 96.2] & 82.1 [67.7, 96.2] & 84.0 [71.5, 97.2] & 85.2 [73.9, 97.2] \\
40 & 94.8 [90.7, 98.0] & 96.0 [94.1, 97.8] & 95.8 [92.7, 98.3] & 95.5 [92.2, 98.1] \\
53 & 94.5 [91.0, 97.9] & 95.1 [92.1, 98.0] & 95.3 [91.7, 98.6] & 96.3 [93.7, 98.7] \\
\bottomrule
\end{tabular}
\caption{Full Sonnet content-label miss rates over the cross-lingual content pool (Eng-orig $\cap$ Rus-para). Content $N$: L16=1{,}072, L31=2{,}236, L40=2{,}143, L53=1{,}392. Random baselines confirm content selectivity (features rarely fire on controlled-unrelated sentences). 95\% bootstrap CIs ($10{,}000$ resamples), feature-clustered.}
\label{tab:appendix-sonnet-maxvar}
\end{table*}

\begin{table*}[t]
\centering
\small
\setlength{\tabcolsep}{3.5pt}
\begin{tabular}{@{}lcccccccc@{}}
\toprule
 & \multicolumn{2}{c}{Within-language (\%)} & \multicolumn{2}{c}{Serbian orig (\%)} & \multicolumn{2}{c}{Serbian para (\%)} & \multicolumn{2}{c}{Cyr/Lat ratio} \\
\cmidrule(lr){2-3} \cmidrule(lr){4-5} \cmidrule(lr){6-7} \cmidrule(lr){8-9}
Layer & Eng-para & Rus-orig & Sr-Latin & Sr-Cyrillic & Sr-Latin & Sr-Cyrillic & Orig & Para \\
\midrule
16 & 13.6 [11.5, 16.4] & 15.0 [12.3, 18.5] & 19.2 [15.6, 23.7] & 19.8 [16.0, 24.7] & 17.8 [14.1, 22.3] & 19.2 [15.3, 24.2] & 1.03$\times$ & 1.08$\times$ \\
31 &  9.5 [7.6, 11.4]  & 12.8 [10.2, 15.7] & 20.7 [17.9, 24.1] & 22.8 [19.9, 26.2] & 18.1 [15.5, 21.3] & 21.3 [18.5, 24.8] & 1.10$\times$ & 1.18$\times$ \\
40 &  8.3 [6.5, 10.3]  & 15.0 [11.9, 17.9] & 20.2 [16.5, 24.0] & 24.3 [20.3, 28.5] & 19.0 [15.5, 22.6] & 23.4 [19.8, 27.4] & 1.21$\times$ & 1.23$\times$ \\
53 &  7.9 [6.2, 10.0]  & 12.9 [10.2, 15.9] & 22.3 [18.9, 26.3] & 30.3 [26.3, 34.8] & 23.7 [20.4, 27.7] & 31.0 [27.6, 35.1] & 1.36$\times$ & 1.31$\times$ \\
\midrule
 & \multicolumn{4}{c}{Random baselines (\%)} \\
\cmidrule(lr){2-5}
Layer & Eng & Rus & Sr-Latin & Sr-Cyrillic \\
\cmidrule(lr){2-5}
16 & 81.6 [77.3, 87.7] & 80.6 [75.7, 87.1] & 81.5 [75.9, 88.3] & 80.6 [75.0, 87.6] \\
31 & 82.6 [69.6, 94.0] & 82.5 [68.9, 94.1] & 84.2 [72.6, 94.6] & 84.8 [73.6, 94.8] \\
40 & 84.3 [69.6, 95.9] & 85.0 [70.3, 96.0] & 84.5 [69.8, 96.1] & 83.9 [68.4, 95.9] \\
53 & 82.4 [67.4, 95.2] & 84.1 [70.0, 96.1] & 85.6 [72.9, 96.8] & 87.1 [75.4, 97.0] \\
\bottomrule
\end{tabular}
\caption{Full Gemini content-label miss rates over the cross-lingual content pool (Eng-orig $\cap$ Rus-para). Content $N$: L16=1{,}503, L31=2{,}444, L40=2{,}744, L53=2{,}067. Gemini classifies more features as content-claim than Sonnet, yielding larger $N$; the asymmetry pattern is consistent across labelers. 95\% bootstrap CIs ($10{,}000$ resamples), feature-clustered.}
\label{tab:appendix-gemini-maxvar}
\end{table*}

\begin{table*}[t]
\centering
\small
\setlength{\tabcolsep}{3.5pt}
\begin{tabular}{@{}lcccccccc@{}}
\toprule
 & \multicolumn{4}{c}{Sonnet} & \multicolumn{4}{c}{Gemini} \\
\cmidrule(lr){2-5} \cmidrule(lr){6-9}
Layer & $N$ & Sr-Lat & Sr-Cyr & Ratio & $N$ & Sr-Lat & Sr-Cyr & Ratio \\
\midrule
16 & 1{,}072 & 20.8 [15.9, 27.4] & 21.3 [15.8, 28.8] & 1.02$\times$ [0.91, 1.12] & 1{,}503 & 19.2 [15.6, 23.7] & 19.8 [16.0, 24.7] & 1.03$\times$ [0.94, 1.11] \\
31 & 2{,}236 & 20.4 [18.1, 23.4] & 22.6 [20.3, 25.1] & 1.11$\times$ [1.01, 1.19] & 2{,}444 & 20.7 [17.9, 24.1] & 22.8 [19.9, 26.2] & 1.10$\times$ [1.03, 1.18] \\
40 & 2{,}143 & 22.2 [19.9, 24.6] & 25.9 [23.4, 28.6] & 1.17$\times$ [1.09, 1.26] & 2{,}744 & 20.2 [16.5, 24.0] & 24.3 [20.3, 28.5] & 1.21$\times$ [1.13, 1.29] \\
53 & 1{,}392 & 25.6 [22.1, 29.6] & 34.6 [30.7, 38.6] & 1.35$\times$ [1.23, 1.49] & 2{,}067 & 22.3 [18.9, 26.3] & 30.3 [26.3, 34.8] & 1.36$\times$ [1.25, 1.47] \\
\bottomrule
\end{tabular}
\caption{Sonnet vs.\ Gemini labeler comparison over the cross-lingual content pool. Content $N$ is the number of (feature, sentence) pairs classified as content-claim. This count differs between labelers because Sonnet and Gemini describe features differently, producing distinct content-claim subsets from the same underlying activation pool; the converging ratios confirm the signal is robust to labeler choice. Brackets give 95\% bootstrap CIs ($10{,}000$ resamples, feature-clustered).}
\label{tab:labeler-comparison-maxvar}
\end{table*}

\begin{table*}[t]
\centering
\footnotesize
\begin{tabular}{@{}p{0.47\textwidth}p{0.47\textwidth}@{}}
\toprule
\multicolumn{2}{@{}l}{\textbf{Misses Sr-Cyrillic} (fires English, Russian, and Sr-Latin)} \\
\midrule
\textbf{L16 / F686} \newline \textbf{Auto-interp:} Sonnet: ``historical events'' | Gemini: ``negative historical events'' \newline \textbf{Sentence:} ``The photographer was transported to Ronald Reagan UCLA Medical Center, where he subsequently died.''
&
\textbf{L16 / F12470} \newline \textbf{Auto-interp:} Sonnet: ``airport'' | Gemini: ``airport security'' \newline \textbf{Sentence:} ``Today, Timbuktu is an impoverished town, although its reputation makes it a tourist attraction, and it has an airport.''
\\[6pt]
\textbf{L16 / F2851} \newline \textbf{Auto-interp:} Sonnet: ``damage'' | Gemini: ``damage, accidents, and injuries'' \newline \textbf{Sentence:} ``Water is spilling over the levee in a section 100 feet wide.''
&
\textbf{L31 / F2733} \newline \textbf{Auto-interp:} Sonnet: ``corporate business'' | Gemini: ``corporate strategies and business news'' \newline \textbf{Sentence:} ``Last week, METI announced that Apple had informed it of 34 additional overheating incidents, which the company called `non-serious.'\,''
\\[6pt]
\textbf{L31 / F11478} \newline \textbf{Auto-interp:} Sonnet: ``emergency preparedness'' | Gemini: ``emergency preparedness and response'' \newline \textbf{Sentence:} ``Negotiators tried to rectify the situation, but the prisoners' demands are not clear.''
&
\textbf{L31 / F8176} \newline \textbf{Auto-interp:} Sonnet: ``tourism'' | Gemini: ``travel and tourism'' \newline \textbf{Sentence:} ``For example visiting castles in the Loire Valley, the Rhine valley or taking a cruise to interesting cites on the Danube or boating along the Erie Canal.''
\\[6pt]
\textbf{L31 / F1773} \newline \textbf{Auto-interp:} Sonnet: ``political'' | Gemini: ``political figures and roles'' \newline \textbf{Sentence:} ``Throughout 1960s, Brzezinski worked for John F. Kennedy as his advisor and then the Lyndon B. Johnson administration.''
&
\textbf{L40 / F4378} \newline \textbf{Auto-interp:} Sonnet: ``disaster recovery'' | Gemini: ``disaster, damage, and recovery'' \newline \textbf{Sentence:} ``The Ninth Ward, which saw flooding as high as 20 feet during Hurricane Katrina, is currently in waist-high water as the nearby levee was overtopped.''
\\[6pt]
\textbf{L40 / F7069} \newline \textbf{Auto-interp:} Sonnet: ``musical instruments'' | Gemini: ``musical instruments'' \newline \textbf{Sentence:} ``Up means you should start at the tip and push the bow, and down means you should start at the frog (which is where your hand is holding the bow) and pull the bow.''
&
\textbf{L40 / F4430} \newline \textbf{Auto-interp:} Sonnet: ``mice'' | Gemini: ``animals in studies'' \newline \textbf{Sentence:} ``\,`We now have 4-month-old mice that are non-diabetic that used to be diabetic,' he added.''
\\[6pt]
\textbf{L40 / F14448} \newline \textbf{Auto-interp:} Sonnet: ``border'' | Gemini: ``border'' \newline \textbf{Sentence:} ``The Oyapock River Bridge is a cable-stayed bridge. It spans the Oyapock River to link the cities of Oiapoque in Brazil and Saint-Georges de l'Oyapock in French Guiana.''
&
\textbf{L53 / F12525} \newline \textbf{Auto-interp:} Sonnet: ``male names'' | Gemini: ``male names'' \newline \textbf{Sentence:} ``Edgar Veguilla received arm and jaw wounds while Kristoffer Schneider was left requiring reconstructive surgery for his face.''
\\[6pt]
\textbf{L53 / F12994} \newline \textbf{Auto-interp:} Sonnet: ``policy and government'' | Gemini: ``policy and government institutions'' \newline \textbf{Sentence:} ``The money could go toward flood-proof houses, better water management, and crop diversification.''
&
\textbf{L53 / F3443} \newline \textbf{Auto-interp:} Sonnet: ``travel'' | Gemini: ``travel and movement'' \newline \textbf{Sentence:} ``If the country you will be visiting becomes subject to a travel advisory, your travel health insurance or your trip cancellation insurance may be affected.''
\\[6pt]
\textbf{L53 / F7923} \newline \textbf{Auto-interp:} Sonnet: ``Muslim'' | Gemini: ``Muslim discourse and related concepts'' \newline \textbf{Sentence:} ``Muhammad was deeply interested in matters beyond this mundane life. He used to frequent a cave that became known as `Hira' on the Mountain of `Noor' (light) for contemplation.''
&
\\
\midrule
\multicolumn{2}{@{}l}{\textbf{Misses Sr-Latin} (fires English, Russian, and Sr-Cyrillic)} \\
\midrule
\textbf{L16 / F2475} \newline \textbf{Auto-interp:} Sonnet: ``Banking and financial institutions'' | Gemini: ``banking and financial institutions'' \newline \textbf{Sentence:} ``Money can be exchanged at the only bank in the islands which is located in Stanley across from the FIC West store.''
&
\textbf{L31 / F1938} \newline \textbf{Auto-interp:} Sonnet: ``news and politics'' | Gemini: ``news and political commentary'' \newline \textbf{Sentence:} ``Suspected cases of H5N1 in Croatia and Denmark remain unconfirmed.''
\\[6pt]
\textbf{L40 / F2014} \newline \textbf{Auto-interp:} Sonnet: ``ceremony'' | Gemini: ``ceremony'' \newline \textbf{Sentence:} ``Several large television screens were installed in various places in Rome to let the people watch the ceremony.''
&
\textbf{L53 / F1527} \newline \textbf{Auto-interp:} Sonnet: ``access to resources'' | Gemini: ``adequate access to resources'' \newline \textbf{Sentence:} ``The cities of Harappa and Mohenjo-daro had a flush toilet in almost every house, attached to a sophisticated sewage system.''
\\[6pt]
\textbf{L53 / F4334} \newline \textbf{Auto-interp:} Sonnet: ``Government and governmental institutions'' | Gemini: ``government'' \newline \textbf{Sentence:} ``The Articles required unanimous consent from all the states before they could be amended and states took the central government so lightly that their representatives were often absent.''
&
\\
\bottomrule
\end{tabular}
\caption{Extended examples of content-labeled features whose auto-interp labels fail on one Serbian script. All verifiable on Neuronpedia (\texttt{gemma-3-27b}, SAE \texttt{\{layer\}-gemmascope-2-res-16k}).}
\label{tab:vignettes-appendix}
\end{table*}

\begin{table*}[h]
\centering
\small
\setlength{\tabcolsep}{6pt}
\begin{tabular}{@{}lrccc@{}}
\toprule
Topic & $N$ & Sr-Latin (\%) & Sr-Cyrillic (\%) & Cyr/Lat ratio \\
\midrule
Travel \& Tourism        & 1{,}998 & 23.1 [20.8, 25.6] & 27.3 [24.8, 30.0] & 1.18$\times$ [1.11, 1.27] \\
Science \& Technology    & 1{,}746 & 20.8 [18.5, 23.4] & 23.8 [21.3, 26.5] & 1.14$\times$ [1.04, 1.25] \\
History \& Geography     &   672   & 19.8 [16.5, 23.4] & 24.3 [20.6, 28.2] & 1.23$\times$ [1.09, 1.38] \\
Politics \& Law          &   859   & 21.2 [17.9, 24.8] & 25.7 [22.1, 29.7] & 1.21$\times$ [1.11, 1.33] \\
Culture \& Entertainment &   478   & 26.8 [22.3, 31.7] & 30.8 [26.2, 35.7] & 1.15$\times$ [1.04, 1.28] \\
Other (small, merged)    & 1{,}090 & 22.4 [19.3, 25.7] & 25.5 [22.4, 28.8] & 1.14$\times$ [1.04, 1.25] \\
\bottomrule
\end{tabular}
\caption{Cyrillic/Latin miss-rate asymmetry decomposed by FLORES+ topic on the 27B Sonnet content pool. Each sentence is assigned to its first topic tag. $N$ counts (feature, sentence) units pooled across layers 16, 31, 40, and 53. 95\% feature-clustered bootstrap CIs in brackets (10{,}000 resamples; paired draws for ratios). The Cyrillic disadvantage holds within every topic bucket — every ratio CI excludes 1.0 — including entity-light topics (Science \& Technology, Travel \& Tourism), indicating the effect is not driven by named entities. The merged Other row aggregates Sports, Education \& Social Science, Disasters \& Accidents, and Other, which are individually too small for stable per-layer bootstraps.}
\label{tab:per-topic-asymmetry}
\end{table*}

\begin{table*}[t]
\centering
\small
\setlength{\tabcolsep}{6pt}
\begin{tabular}{@{}lrccc@{}}
\toprule
Topic & $N$ & Sr-Latin (\%) & Sr-Cyrillic (\%) & Cyr/Lat ratio \\
\midrule
Travel \& Tourism        & 2{,}116 & 22.7 [20.5, 25.2] & 27.2 [24.7, 29.9] & 1.20$\times$ [1.12, 1.28] \\
Science \& Technology    & 1{,}790 & 20.9 [18.5, 23.4] & 23.8 [21.2, 26.5] & 1.14$\times$ [1.04, 1.24] \\
History \& Geography     & 1{,}081 & 19.1 [16.4, 21.9] & 23.1 [20.1, 26.3] & 1.21$\times$ [1.11, 1.33] \\
Politics \& Law          &   859   & 21.2 [17.9, 24.6] & 25.7 [22.1, 29.5] & 1.21$\times$ [1.11, 1.33] \\
Culture \& Entertainment &   572   & 26.4 [22.4, 30.7] & 29.5 [25.4, 33.8] & 1.12$\times$ [1.02, 1.24] \\
Other (small, merged)    & 1{,}197 & 22.0 [19.1, 25.2] & 25.8 [22.8, 29.0] & 1.17$\times$ [1.07, 1.29] \\
\bottomrule
\end{tabular}
\caption{Cyrillic/Latin miss-rate asymmetry decomposed by FLORES+ topic on the 27B Sonnet content pool, Multi-Tag mode: each sentence contributes to \emph{every} topic bucket its tags map to (so multi-tagged sentences appear in more than one row). $N$ counts (feature, sentence) units pooled across layers 16, 31, 40, and 53. 95\% feature-clustered bootstrap CIs in brackets (10{,}000 resamples; paired draws for ratios). The Cyrillic disadvantage holds within every topic — every ratio CI excludes 1.0 — confirming that the asymmetry is robust to how multi-tagged sentences are assigned to buckets.}
\label{tab:per-topic-asymmetry-multi}
\end{table*}

\begin{table*}[t]
\centering
\small
\setlength{\tabcolsep}{4pt}
\begin{tabular}{@{}llrrccc@{}}
\toprule
Layer & View & Feats & Pairs & Miss Sr-Lat (95\% CI) & Miss Sr-Cyr (95\% CI) & Ratio (95\% CI) \\
\midrule
16 & unfiltered      &   437 & 4{,}777 & 14.42\% [11.84, 17.36] & 14.78\% [12.09, 18.05] & 1.02$\times$ [0.96, 1.08] \\
16 & Sonnet-content  &   287 & 1{,}072 & 20.80\% [15.92, 27.27] & 21.27\% [15.77, 28.89] & 1.02$\times$ [0.91, 1.12] \\
\midrule
31 & unfiltered      &   867 & 4{,}422 & 20.53\% [17.91, 23.64] & 22.89\% [19.86, 26.15] & 1.11$\times$ [1.05, 1.18] \\
31 & Sonnet-content  &   648 & 2{,}236 & 20.44\% [18.12, 23.36] & 22.58\% [20.35, 25.08] & 1.11$\times$ [1.01, 1.20] \\
\midrule
40 & unfiltered      & 1{,}068 & 4{,}182 & 21.38\% [18.44, 24.65] & 26.37\% [22.86, 29.99] & 1.23$\times$ [1.16, 1.31] \\
40 & Sonnet-content  &   796 & 2{,}143 & 22.17\% [19.92, 24.76] & 25.94\% [23.32, 28.64] & 1.17$\times$ [1.09, 1.26] \\
\midrule
53 & unfiltered      &   923 & 2{,}880 & 24.34\% [21.19, 27.81] & 31.22\% [27.93, 34.85] & 1.28$\times$ [1.20, 1.37] \\
53 & Sonnet-content  &   555 & 1{,}392 & 25.65\% [22.11, 29.54] & 34.55\% [30.76, 38.55] & {1.35$\times$ [1.23, 1.49]} \\
\bottomrule
\end{tabular}
\caption{Gemma~3 27B Serbian Cyrillic miss-rate ratio with feature-clustered bootstrap CIs (10{,}000 resamples, seed 42). For each layer we report two nested views of the same feature pool. \textbf{Unfiltered} contains every (feature, sentence) pair whose feature fires on both the English original and the Russian paraphrase, with no label-based filtering. \textbf{Sonnet-content} is the strict subset of that pool whose features were classified as \emph{content-claim} by the Sonnet labeler (e.g., at layer 16, 287 of the 437 unfiltered features); this is the pool used in the main analyses throughout the paper. Miss rate = fraction of pool pairs where the feature does \emph{not} fire on the named condition; Ratio = Sr-Cyr miss / Sr-Lat miss. The two views give overlapping ratios at every layer and Layer 16 is statistically indistinguishable from parity; layers 31, 40, and 53 each show a significant Sr-Cyr penalty. }
\label{tab:27b_ratio_featureCI}
\end{table*}

\begin{table}[t]
\centering
\small
\begin{tabular}{lccc}
\toprule
Condition & Accepted (\%) & Corrected (\%) & Diff.\ (pp) \\
\midrule
Eng-para$^{\dagger}$   & 9.6  & 10.3 & $-0.7$ [$-4.6$, $+2.7$] \\
Sr-Lat-para            & 19.4 & 22.3 & $-2.8$ [$-4.9$, $-0.9$] \\
Sr-Cyr-para            & 23.9 & 27.2 & $-3.4$ [$-5.6$, $-1.2$] \\
Rus-orig$^{\dagger}$ (anchor) & 15.3 & 13.1 & $+2.3$ [$-0.5$, $+4.9$] \\
\bottomrule
\end{tabular}
\caption{Miss rates split by validation outcome: sentences whose paraphrase was
accepted as-is vs.\ flagged and corrected (Gemma-3-27B, Sonnet content pool,
pooled over layers 16, 31, 40, 53). Diff.\ = accepted $-$ corrected, with 95\%
feature-clustered bootstrap CIs (10,000 resamples). Corrected sentences: English
12, Serbian 88, Russian 28; $^{\dagger}$fewer than 30 corrected sentences. The
Russian paraphrase is a selection anchor, so the Russian original is split by the
paraphrase's correction status. Sentence-clustered CIs for Serbian include zero
(Sr-Cyr [$-7.8$, $+0.7$], Sr-Lat [$-6.9$, $+1.0$]).}
\label{tab:valsplit}
\end{table}

\subsection{Human Verification of Missed Features}
\label{app:humanverif}
We manually inspected a stratified sample of 60 missed (feature, sentence) units
from the Gemma-3-27B Sonnet content pool: 15 per layer (16, 31, 40, 53), split
evenly between Serbian Cyrillic and Serbian Latin misses, with one unit per
feature. For each unit, the author judged (Q1) whether the Sonnet label describes
the feature's top-activating examples on Neuronpedia, and (Q2) whether the
labeled concept is present in the sentence the feature missed. Each question was
answered Yes, Partly, or No. Annotation was done by a single annotator with the
layer and missed script visible, so we report no inter-annotator agreement.
Table~\ref{tab:humanverif} reports the results.

\begin{table}[t]
\centering
\small
\begin{tabular}{lccccc}
\toprule
 & & \multicolumn{2}{c}{Q1: label accurate} & \multicolumn{2}{c}{Q2: concept present} \\
\cmidrule(lr){3-4}\cmidrule(lr){5-6}
Layer & $n$ & Yes & Yes+Partly & Yes & Yes+Partly \\
\midrule
16  & 15 & 93\% & 100\% & 40\% & 60\% \\
31  & 15 & 80\% & 80\%  & 40\% & 53\% \\
40  & 15 & 80\% & 93\%  & 53\% & 67\% \\
53  & 15 & 80\% & 93\%  & 33\% & 53\% \\
\midrule
All & 60 & 83\% [72, 91] & 92\% [82, 96] & 42\% [30, 54] & 58\% [46, 70] \\
\bottomrule
\end{tabular}
\caption{Human verification of 60 sampled missed units. Q1: the Sonnet label
describes the feature's top-activating examples. Q2: the labeled concept is
present in the sentence the feature missed. Brackets give Wilson 95\% CIs;
per-layer differences are not significant.}
\label{tab:humanverif}
\end{table}

The labels are locally accurate: 92\% describe their top-activating examples at
least partly. The labeled concept, however, is present in the missed sentence in
only 58\% of units (42\% strictly). In the remainder, the feature fires on the English and Russian versions of a
sentence that does not contain its labeled concept, indicating that the
Neuronpedia label captures only part of what drives the feature. Both cases
point to the same limitation of auto-interpretation labels: a label accurate on
a feature's top-activating examples does not reliably describe the feature's
behavior on other inputs.

\clearpage
\section{Leg 2 Across Model Scale}
\label{app:leg2_scale}

We replicate the Leg~2 auto-interpretation evaluation on Gemma-3-1B and
Gemma-3-12B. The core pattern holds at both scales
(Tables~\ref{tab:leg2_12b} and~\ref{tab:leg2_1b}): Serbian miss rates
exceed within-language floors, the Cyrillic/Latin asymmetry grows with depth
(reaching $1.18\times$ at 1B and $1.33\times$ at 12B), and miss rates are
ordered by estimated training coverage.

\begin{table}[t]
\centering
\small
\setlength{\tabcolsep}{4pt}

\begin{subtable}[t]{\columnwidth}
\centering
\begin{tabular}{@{}cccc@{}}
\toprule
Layer & Eng-para (\%) & Rus-orig (\%) & Serbian avg (\%) \\
\midrule
12 & 15.5 [9.9, 22.2]  & 19.0 [13.6, 25.5] & 23.2 [17.5, 30.0] \\
24 & 11.7 [8.4, 17.0]  & 13.9 [9.9, 20.8]  & 20.4 [16.1, 28.1] \\
31 &  8.5 [7.0, 10.2]  & 15.2 [13.1, 17.5] & 26.0 [23.0, 29.2] \\
41 & 10.0 [7.1, 13.5]  & 13.6 [10.3, 17.6] & 32.6 [30.0, 36.1] \\
\bottomrule
\end{tabular}
\caption{Within- vs.\ cross-language miss rates.}
\label{tab:leg2_12b_a}
\end{subtable}

\vspace{0.8em}

\begin{subtable}[t]{\columnwidth}
\centering
\begin{tabular}{@{}cccc@{}}
\toprule
Layer & Sr-Latin (\%) & Sr-Cyrillic (\%) & Cyr/Lat \\
\midrule
12 & 22.6 [16.6, 29.7] & 23.8 [17.6, 31.1] & 1.05$\times$ [0.86, 1.30] \\
24 & 20.0 [15.6, 27.6] & 20.9 [16.4, 28.9] & 1.04$\times$ [0.96, 1.14] \\
31 & 24.1 [21.0, 27.5] & 27.8 [24.7, 31.3] & 1.16$\times$ [1.07, 1.25] \\
41 & 27.9 [25.3, 30.6] & 37.3 [33.4, 42.3] & 1.33$\times$ [1.20, 1.50] \\
\bottomrule
\end{tabular}
\caption{Serbian script asymmetry.}
\label{tab:leg2_12b_b}
\end{subtable}

\caption{Content-feature miss rates for Gemma-3-12B (Gemini labels).
95\% bootstrap CIs in brackets (10{,}000 resamples), feature-clustered.}
\label{tab:leg2_12b}
\end{table}

\begin{table}[t]
\centering
\small
\setlength{\tabcolsep}{4pt}

\begin{subtable}[t]{\columnwidth}
\centering
\begin{tabular}{@{}cccc@{}}
\toprule
Layer & Eng-para (\%) & Rus-orig (\%) & Serbian avg (\%) \\
\midrule
 7 &  9.0 [5.0, 15.8]  &  8.5 [4.2, 18.6]  & 13.8 [8.3, 27.3]  \\
13 & 15.2 [12.8, 17.8] & 21.5 [18.6, 24.8] & 35.6 [31.3, 40.2] \\
17 &  7.1 [4.4, 11.1]  & 13.4 [9.5, 18.6]  & 36.8 [25.1, 52.0] \\
22 &  5.9 [3.8, 8.4]   & 14.1 [11.8, 17.6] & 53.8 [42.8, 67.3] \\
\bottomrule
\end{tabular}
\caption{Within- vs.\ cross-language miss rates.}
\label{tab:leg2_1b_a}
\end{subtable}

\vspace{0.8em}

\begin{subtable}[t]{\columnwidth}
\centering
\begin{tabular}{@{}cccc@{}}
\toprule
Layer & Sr-Latin (\%) & Sr-Cyrillic (\%) & Cyr/Lat \\
\midrule
 7 & 13.8 [8.6, 26.0]  & 13.8 [8.0, 28.7]  & 1.00$\times$ [0.85, 1.15] \\
13 & 35.1 [31.0, 39.5] & 36.1 [31.2, 41.4] & 1.03$\times$ [0.95, 1.11] \\
17 & 34.5 [23.7, 48.9] & 39.0 [26.5, 55.3] & 1.13$\times$ [1.04, 1.23] \\
22 & 49.4 [40.0, 62.6] & 58.2 [44.2, 73.9] & 1.18$\times$ [1.00, 1.42] \\
\bottomrule
\end{tabular}
\caption{Serbian script asymmetry.}
\label{tab:leg2_1b_b}
\end{subtable}

\caption{Content-feature miss rates for Gemma-3-1B (Gemini labels).
95\% bootstrap CIs in brackets (10{,}000 resamples), feature-clustered.}
\label{tab:leg2_1b}
\end{table}

\end{document}